%% file: iclr2027_conference.tex
\documentclass{article} 
\usepackage{iclr2027_conference,times}
\usepackage{booktabs}
\input{math_commands.tex}

\usepackage{graphicx,makecell,tabularx}
\usepackage{hyperref}
\usepackage{url}
\usepackage{placeins}
\usepackage{multirow}
\usepackage{amsmath}
\usepackage{algorithm,algpseudocode}
\usepackage{subfigure}
\usepackage{caption}   
\title{Online World-Model-Boosted Planning for Multi-Agent Quadruped Soccer}

\author{Hanzhong~Zhong$^*$, Dongjie~Zhu$^*$, Yu~Chen,  \& Xiang~Li$^\dagger$ 
}

\iclrfinalcopy 
\begin{document}

\maketitle

\begin{abstract}

Learning coordinated behaviors in multi-agent systems remains challenging due to delayed scoring events, strong inter-agent coupling, and large joint action spaces, which make model-free exploration highly sample-inefficient. We study this problem in multi-quadruped robot soccer, which is a representative and commonly-used benchmark. Specifically, each robot selects among pretrained repositioning, dribbling, and kicking skills together with continuous command parameters while competing against mobile opponents through self-play. We propose an online model-based multi-agent reinforcement learning framework in which model predictive control (MPC) method serves as a continually improving training-time teacher for decentralized high-level policies. A skill-level world model is learned online from self-play transitions and used by an MPC to optimize hybrid skill selection and continuous skill parameters. To compensate for the limited planning horizon, a learned centralized value function provides long-term guidance during action evaluation. The resulting MPC decisions are subsequently distilled into decentralized high-level policies for efficient execution. This creates a closed-loop learning process in which evolving policies generate increasingly relevant data for world-model learning, while the improving model provides progressively stronger planning targets for policy optimization. Experiments against model-free Multi-Agent Proximal Policy Optimization (MAPPO) and ablations show that the proposed method improves environment sample efficiency and overall policy performance. The project website is available at \url{https://anonymous.4open.science/w/skills2strategy}.


\end{abstract}

\section{Introduction}
Robot soccer provides a testbed for embodied multi-agent intelligence, linking low-level physical skills with team-level cooperation and competition. Success requires robots to translate tactical decisions into coordinated physical interactions with teammates, opponents, and the ball, making it a useful setting
for studying how motor skills support collective strategies.
Reinforcement learning has enabled robust low-level quadruped skills \citep{10160325,10802022,zhu2026dynamic}, while hierarchical systems combine pretrained skills with high-level MAPPO and self-play for cooperative and competitive soccer \citep{pmlrv305su25a}. However, learning effective team strategies remains interaction-intensive: discrete skill selections and
continuous commands create a large joint action space, compounded by delayed scoring events and coupled robot--ball interactions
\citep{yang2024sample}.

Existing approaches face several limitations in this setting. Model-free self-play relies heavily on environment interaction to discover effective strategies. Learned models can support action evaluation, but models trained on fixed datasets may become inaccurate as self-play changes the state-action distribution. Model-based planning also faces a horizon trade-off: short rollouts can overlook delayed rewards, whereas longer rollouts accumulate prediction errors and increase computational cost. Moreover, centralized planners rely on joint information that is unavailable to decentralized actors, limiting their direct use at deployment.

\begin{figure}[!tp]
\centering
\includegraphics[width=0.8\linewidth]{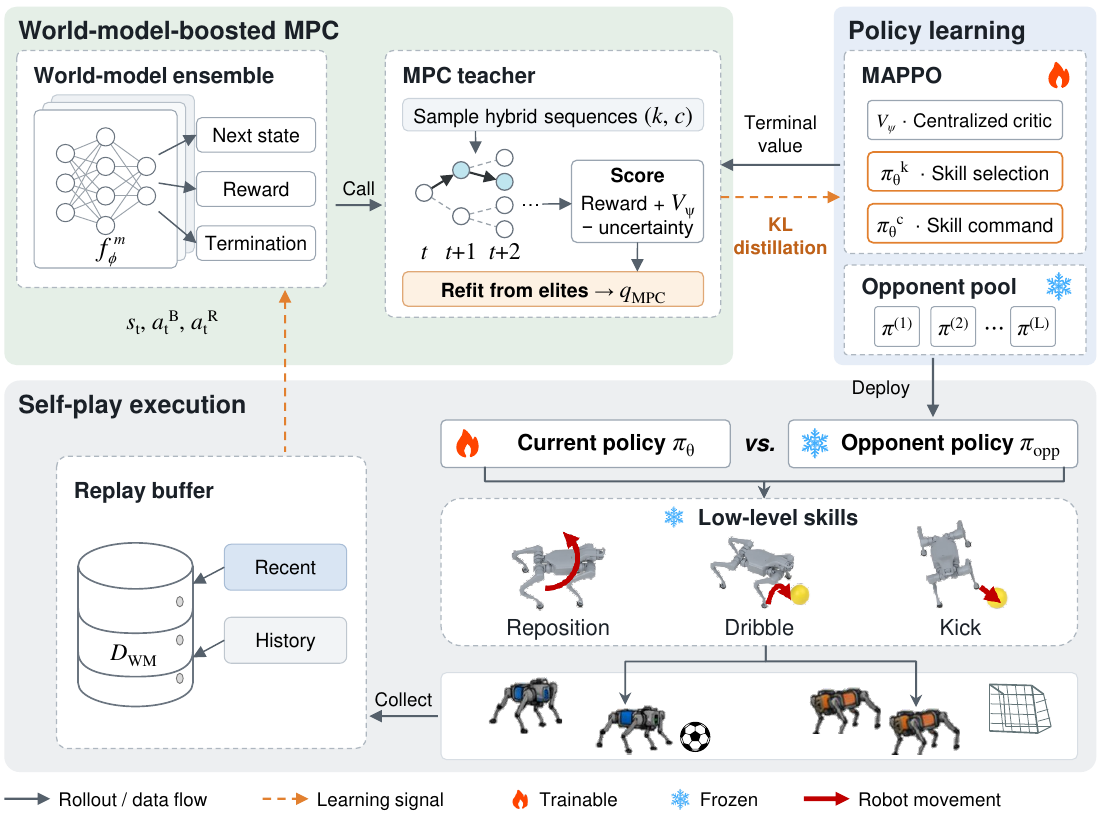}
\caption{Overview of the proposed framework. Self-play trajectories are used to continually train the skill-level world model, which enables MPC to provide improved high-level action targets for the decentralized policies.}
\label{fig:overview}
\end{figure}

We address these challenges with an online world-model-boosted Multi-Agent Proximal Policy Optimization (MAPPO) framework. As illustrated in Figure~\ref{fig:overview}, policy-generated
self-play trajectories support world-model learning, Model Predictive Control (MPC) supervision, and on-policy updates. Historical checkpoints
supply frozen opponents for both execution and planning, while the pretrained low-level skills remain fixed.

First, a probabilistic ensemble world model predicts skill-level
transitions, rewards, and termination from the global state and
both teams' joint actions, separating dynamics prediction from
opponent action generation. Training on recent and historical
self-play data supports adaptation while retaining past experience.

Second, an uncertainty-aware MPC teacher jointly optimizes the controlled team's skill selections and continuous commands using a hybrid cross-entropy method (CEM). Candidate sequences are scored using predicted rewards, a centralized critic's
terminal value, and ensemble disagreement. The terminal value estimates returns beyond short rollouts, while the disagreement penalty discourages reliance on uncertain predictions. Building on value-guided MPC \citep{hansen2024tdmpc2,romero2026actorcritic},
the teacher plans over multi-agent hybrid actions with opponent responses generated by the same frozen policy used in self-play.

Third, MPC transfers privileged planning guidance into
decentralized actors through auxiliary KL distillation of its
optimized first-step action distributions. Actors generate
all environment actions, and MAPPO returns and advantages are
computed from simulator transitions. Thus, unlike methods using
imagined experience for policy-gradient updates
\citep{venugopal2023mabl,shi2025gawm}, our model supplies
action-distribution targets for policy learning.
Deployment requires only the decentralized actors and
pretrained low-level skills.

Our contribution is joint hybrid-action planning and decentralized
policy learning in competitive self-play. We build on established
modeling, MPC, and distillation components. Our contributions are:
\begin{enumerate}
\item A training-time joint search over teammates' skill choices and
continuous commands, addressing coupled action evaluation while keeping
environment interaction and PPO targets policy-generated.
\item Transfer of joint-planning marginals into local recurrent actors,
with direct comparisons against students of independent planners to
measure what coordination survives decentralized execution.
\item Controlled evidence separating joint search, replay coverage, and
the feasibility prior, together with candidate-ranking and compute-cost
diagnostics that delimit the benefits of model-guided learning.
\end{enumerate}

\section{Related Work}
\label{sec:related_work}

Research on robot soccer spans low-level locomotion and ball
manipulation, hierarchical decision-making, and multi-agent
self-play.
Deep reinforcement learning has enabled quadruped robots to acquire
dynamic locomotion and ball-manipulation skills that are difficult to
engineer manually, including robust dribbling and hierarchical
ball-control behaviors \citep{10160325,zhu2026dynamic}. These works
primarily address physical skill execution rather than multi-robot coordination. More recent systems combine learned skills with high-level Multi-Agent Reinforcement Learning (MARL) to achieve cooperative and competitive soccer \citep{pmlrv305su25a,tirumala2025robot}. Large-scale league training has also produced full robotic-football strategies in simulation \citep{smit2023football}, while MQE provides environments for studying
multi-agent quadruped tasks but relies mainly on a built-in locomotion controller \citep{xiong2024mqe}. MAPPO provides a strong centralized-training and decentralized-execution
baseline, while historical self-play, PSRO, and league training expose agents to diverse and increasingly capable opponents
\citep{yu2022surprising,lanctot2017psro,vinyals2019alphastar}. In quadruped soccer, \citet{pmlrv305su25a} combine MAPPO with fictitious self-play but discretize high-level command directions to simplify learning. Such model-free, on-policy training can require substantial interaction before coordinated tactical behavior emerges \citep{yang2024sample}. We instead retain continuous skill commands and use an online learned model and MPC teacher to reduce the exploration
burden of high-level self-play.


Model-based reinforcement learning improves interaction efficiency by learning predictive dynamics for planning or imagined policy updates \citep{moerland2023mbrlsurvey}. Probabilistic ensembles and CEM support uncertainty-aware trajectory optimization, while short model rollouts
reduce compounding prediction error \citep{chua2018deep,janner2019trust}. Modern world models further demonstrate that learned dynamics can support policy learning across diverse continuous-control tasks \citep{hafner2025dreamer}. Extending these ideas to MARL is difficult because transitions depend on coupled actions and the behavior of other agents changes during training. Existing methods address this through interaction-aware prediction, opponent modeling, structured latent dynamics, or pessimistic objectives \citep{krupnik2020multi,wu2023models,zhang2021aorpo,
venugopal2023mabl,wen2026mapmbrl}. MAMBPO, MABL, and GAWM primarily use
learned models to generate imagined experience for policy optimization
\citep{willemsen2021mambpo,venugopal2023mabl,shi2025gawm}. This can improve sample efficiency, but directly training on synthetic trajectories exposes the policy to model bias, particularly as opponents and visitation distributions evolve. Our model predicts skill-level transitions conditioned on both teams' hybrid actions. It supports short-horizon planning and action-distribution supervision, while MAPPO returns and advantages are computed from simulator experience.


Learned MPC exploits a predictive model through receding-horizon optimization, but short planning horizons may lead to myopic decisions. TD-MPC and TD-MPC2 address this issue by combining short-horizon planning with a learned terminal value, and related actor--critic MPC methods use learned value estimates to provide long-term guidance
\citep{hansen2022tdmpc,hansen2024tdmpc2,romero2026actorcritic}. We adopt this general principle but plan over the discrete skill selections and continuous commands of multiple robots. The centralized MAPPO critic is
reused as the terminal value function, while ensemble disagreement discourages plans that rely on uncertain world-model predictions. Policy distillation and interactive imitation learning transfer teacher guidance into compact policies
\citep{rusu2015policy,ross2011reduction}. Recent methods similarly use privileged or sampling-based MPC teachers for robot learning \citep{amiri2026pripg,schuck2026learning}. Building on these advances, our framework integrates online world-model learning, uncertainty-aware hybrid planning, and policy distillation for competitive multi-agent quadruped soccer.
The model provides planning-based supervision for decentralized
MAPPO actors, while policy-gradient updates use simulator
experience and deployment requires neither the model nor MPC.

\section{Method}
\label{sec:method}


\subsection{Hierarchical Control and Online World Modeling}
\label{sec:problem_formulation}
We model multi-robot soccer between teams $\mathcal{B}$ and $\mathcal{R}$ as a two-team Markov game, with $N=2$ robots per team. Individual actors have partial observations, while training uses global state information.
At high-level decision step $t$, the global state is denoted by $s_t\in\mathcal{S}$, which includes robot poses, velocities, joint positions and joint velocities at skill boundaries, and the ball state. Each robot $i$ receives a local observation $o_t^i=\mathcal{O}_i(s_t)$ including proprioception and robot identity indicators.

All robots interact with the environment through the following library of pretrained low-level skills:
\begin{equation}
\mathcal{K}
=
\{
\texttt{reposition},
\texttt{dribble},
\texttt{kick}
\}.
\end{equation}

For each robot $i$ in team $X\in\{\mathcal B,\mathcal R\}$, the high-level policy outputs a hybrid action consisting of a discrete skill and its corresponding continuous command parameters:
\begin{equation}
a_t^{X,i}
=
\left(k_t^{X,i},c_t^{X,i}\right),
\qquad
k_t^{X,i}\in\mathcal{K},
\quad
c_t^{X,i}\in\mathbb{R}^{3},
\end{equation}
where $k_t^{X,i}$ and $c_t^{X,i}$ denote the selected skill and its associated continuous parameters, respectively. The three command components and their ranges are defined for each skill in Table \ref{tab:low_level_skills} of Appendix~\ref{apdx:low level skills}. The joint action of team $X$ at time step $t$ is given by $
a_t^X
=
\left(a_t^{X,1},\ldots,a_t^{X,N}\right),
$ and the complete joint action of both teams is defined as
$\bm a_t=(a_t^{\mathcal{B}},a_t^{\mathcal{R}})$.


Executing the selected skills over one high-level interval
produces the next state $s_{t+1}$ and reward $r_t$.
For example, dribbling changes both the robot's position and
the ball's motion, while an opponent's simultaneous movement
may lead to an interception. The world model predicts these
joint interaction outcomes at the skill timescale.
High- and low-level reward definitions are provided in
Tables~\ref{tab: high level reward} and Table~\ref{tab: low level reward} of Appendix~\ref{apdx: reward setting}, respectively.


Rather than predicting complicated joint-level quadruped dynamics, the world model proposed in this work aims to predict the state transition at the high-level skill timescale.
\begin{equation}
p_{\phi}(s_{t+1},r_t,d_t\mid s_t,a_t^{\mathcal{B}},a_t^{\mathcal{R}}).
\end{equation}
where $d_t$ indicates whether the episode terminates after the transition at step $t$

We implement this conditional distribution using an ensemble of $M$
probabilistic models. Each model predicts a residual distribution for
the continuous state variables, together with the reward distribution
and the termination probability $p_d$:
\begin{equation}
f_{\phi}^m
\left(
s_t,a_t^{\mathcal{B}},a_t^{\mathcal{R}}
\right)
=
\left(
\boldsymbol{\mu}_{\Delta s}^{m},
\log \boldsymbol{\sigma}_{\Delta s}^{2,m},
\mu_r^m,
\log \sigma_r^{2,m},
p_d^m
\right).
\end{equation}
where $m \in \{1,\ldots,M\}$ indexes the ensemble members,
  and $\phi$ denotes the trainable parameters of the model; $\boldsymbol{\mu}_{\Delta s}^{m}$ and $\mu_r^m$ are
  the predicted means of the state residual and reward, respectively;
  $\boldsymbol{\sigma}_{\Delta s}^{2,m}$ and $\sigma_r^{2,m}$
  are their predicted variances; and $p_d^m$ is the predicted
  probability of episode termination. Then the predicted continuous state is given by
\begin{equation}
\hat{s}_{t+1}^{m}
=
s_t+\boldsymbol{\mu}_{\Delta s}^{m}.
\end{equation}
The training objective is
\begin{equation}
\mathcal{L}_{\mathrm{WM}}
=
\frac{1}{M}\sum_{m=1}^{M}
\left(
\mathcal{L}_{\mathrm{state}}^{m}
+
\lambda_r\mathcal{L}_{r}^{m}
+
\lambda_d\mathcal{L}_{d}^{m}
\right),
\end{equation}
Here, $\mathcal{L}_{\mathrm{state}}$ denotes the negative
log-likelihood of the predicted state transition,
$\mathcal{L}_{r}$ denotes the Gaussian negative log-likelihood of the reward, and $\mathcal{L}_{d}$ denotes the binary cross-entropy loss for termination prediction.

The model is updated periodically throughout training using recent and
historical self-play transitions. This allows it to adapt to the current
policies while retaining experience from earlier opponents. 
We use historical self-play to reduce short-term non-stationarity. The
current policy is trained against a frozen opponent sampled from a
population of previous checkpoints,
\begin{equation}
\mathcal{P}
=
\left\{
\pi^{(1)},\ldots,\pi^{(L)}
\right\},
\label{eq:op pool}
\end{equation}
At the end of each training interval, the new policy may be added to
this population and a new opponent is selected. 

\subsection{Model Predictive Control}
\label{sec:mpc}
 At state $s_t$, MPC searches over the controlled-team action sequence. 
\begin{equation}
\mathbf A_t^\mathcal{B}
=
\{
\mathbf a_{t}^\mathcal{B},\ldots,
\mathbf a_{t+H-1}^\mathcal{B}
\}.
\end{equation}

At each imagined step, the opponent action is sampled from the
frozen opponent policy using the opponent's predicted observation $\mathcal{O}_{\mathcal{R}}$
\begin{equation}
\hat a_{t+h}^{\mathcal{R}}
\sim
\pi_{\mathrm{opp}}
\left(
\cdot\mid
\mathcal{O}_{\mathcal{R}}(\hat s_{t+h})
\right).
\end{equation}
and imagined transitions are generated according to
\begin{equation}
\hat s_{t+h+1}
\sim
f_\phi^m
\left(
\hat s_{t+h},
\hat{\mathbf a}_{t+h}^\mathcal{B},
\hat{\mathbf a}_{t+h}^\mathcal{R}
\right).
\end{equation}
\begin{figure}[!tp]
\centering
\includegraphics[width=0.8\linewidth]{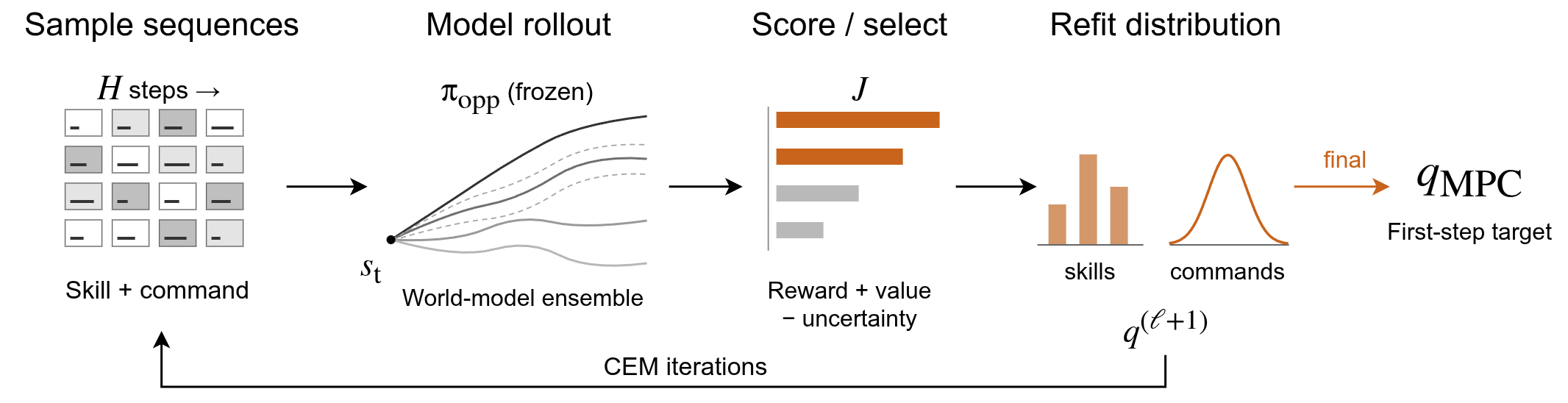}
\caption{Overview of CEM planning within the MPC teacher. Candidate hybrid action sequences are sampled from an evolving distribution and evaluated through ensemble world-model rollouts. Elite sequences are used to refine the sampling distribution over successive CEM iterations. The optimized first-step action distribution provides a distillation target for the decentralized actors.}
\label{fig:cem}
\end{figure}

Since each action contains a discrete skill and continuous parameters, we
use a hybrid CEM as shown in Figure \ref{fig:cem}. For robot $i$ and planning step
$h$, CEM maintains a categorical skill distribution
\begin{equation}
q_{h,i}^{k}(k)
=
\Pr(k_{t+h}^{\mathcal{B},i}=k),
\end{equation}
and a skill-conditioned Gaussian command distribution
\begin{equation}
c_{t+h}^{\mathcal{B},i}\mid k
\sim
\mathcal N 
\left(
\mu_{h,i,k},
\operatorname{diag}(\sigma_{h,i,k}^2)
\right).
\end{equation}
A candidate sequence may assign dribbling to the ball-carrying
robot and repositioning to its teammate, followed by a kick.
The world model evaluates their joint effects under predicted
opponent responses, and the highest-scoring candidates refine
the CEM distributions to obtain $q_{\mathrm{MPC}}$.
For robots far from the ball, sampling is restricted to
repositioning because the dribbling and kicking controllers
were trained for nearby-ball interactions. This restriction
applies only to teacher sampling.

To limit model error, we use a short planning horizon and bootstrap the
remaining return with the centralized MAPPO critic $V_\psi$. For a
candidate trajectory, the termination-aware objective is
\begin{equation}
J
=
\sum_{h=0}^{H-1}\gamma^h b_h
\left(
\hat r_{t+h}
-
\lambda_u U(\hat{s}_{t+h},\hat a^\mathcal{B}_{t+h},\hat a^\mathcal{R}_{t+h})
\right)
+
\gamma^H b_H V_{\psi}(\hat{s}_{t+H}).
\label{eq:mpc_objective}
\end{equation}
Here $b_0=1$ and $b_{h+1}=b_h(1-\hat d_{t+h})$, where
$\hat d_{t+h}\in\{0,1\}$ indicates termination after the imagined
transition. The terminating transition retains its reward, subsequent
rewards and penalties vanish, and bootstrapping is disabled even when
termination occurs on the final rollout step. Each ensemble rollout is
masked separately before scores are averaged. We use the deterministic
rule $\hat d=\mathbf{1}[p_d\geq0.5]$ and stop each terminated rollout
immediately. Terminal values use branch-specific critic
memory and are converted to raw reward units; Appendix~\ref{apdx:implementation}
specifies the recurrent update and observation construction.
The quantity $U(\hat{s}_{t+h},\hat{a}^\mathcal{B}_{t+h},\hat{a}^\mathcal{R}_{t+h})$ measures the disagreement among the ensemble predictions at the corresponding imagined state and action:
\begin{equation}
U(\hat{s},\hat a^{\mathcal B},
\hat a^{\mathcal R})
=
\frac{1}{M}
\sum_{m=1}^{M}
\left(
\hat{s}_{t+1}^{m}
-
\bar{s}_{t+1}
\right)^\top
\mathbf{W}_s
\left(
\hat{s}_{t+1}^{m}
-
\bar{s}_{t+1}
\right),
\qquad
\bar{s}_{t+1}
=
\frac{1}{M}\sum_{j=1}^{M}\hat{s}_{t+1}^{j}.
\end{equation}
where $\mathbf{W}_{s}$ normalizes the state dimensions before
computing the disagreement. For example, dribbling may create a favorable shooting position without yielding a goal within the rollout horizon; the terminal value estimates the subsequent return. A candidate involving
robot-ball contact receives an uncertainty penalty when ensemble
members disagree about the resulting motion.



\subsection{MPC-Guided MAPPO and Online Training}
\label{sec:policy_training}

Environment actions are sampled from the current policy, preserving the
on-policy MAPPO objective. For robot $i$, the PPO ratio is
\begin{equation}
\rho_t^i(\theta)
=
\frac{\pi_{\theta}(a_t^i \mid o_t^i)}
{\pi_{\theta^{\mathrm{old}}}(a_t^i\mid o_t^i)},
\end{equation}
with clipped objective
\begin{equation}
\mathcal J_{\mathrm{PPO}}^i
=
\mathbb E_t
\left[
\min
\left(
\rho_t^i A_t,
\operatorname{clip}
(\rho_t^i,1-\epsilon,1+\epsilon)A_t
\right)
\right].
\end{equation}
MPC is used as an auxiliary teacher rather than as the behavior policy.
For selected states, its skill and command distributions are distilled
into the decentralized actors:
\begin{equation}
\mathcal L_{\mathrm{skill}}^i
=
D_{\mathrm{KL}}
\left(
q_{\mathrm{MPC},i}^{k}
\|
\pi_{\theta}^{k}
(\cdot\mid o_t^i)
\right),
\end{equation}
\begin{equation}
\mathcal L_{\mathrm{cmd}}^i
=
\sum_{k\in\mathcal K}
q_{\mathrm{MPC},i}^{k}(k)
D_{\mathrm{KL}}
\left(
q_{\mathrm{MPC},i}^{c}(\cdot\mid k)
\|
\pi_{\theta}^{c}(\cdot\mid o_t^i,k)
\right).
\end{equation}
where the resulting CEM distributions are denoted by
$q_{\mathrm{MPC},i}^{k}$ for the skill distribution and
$q_{\mathrm{MPC},i}^{c}(\cdot\mid k)$ for the
skill-conditioned command distribution. Joint candidate scoring can
favor complementary actions, although the factorized CEM distribution
and marginal KL targets do not preserve arbitrary joint correlations.
Actors can retain coordination that is predictable from their local
observation histories; exact recovery of a privileged teacher is not
guaranteed. Nevertheless, the teacher--actor comparison in
Appendix~\ref{apdx:controls} shows that the decentralized actor retains
improvements in obstruction and recovery relative to MAPPO; these are
metric-specific comparisons, not an overall coordination-retention score.

The total distillation loss is
\begin{equation}
\mathcal L_{\mathrm{MPC}}
=
\frac{1}{N}
\sum_{i=1}^{N}
\left(
\mathcal L_{\mathrm{skill}}^i
+
\lambda_c\mathcal L_{\mathrm{cmd}}^i
\right),
\end{equation}
and the actor objective is
\begin{equation}
\mathcal L_{\mathrm{actor}}
=
-\mathcal J_{\mathrm{PPO}}
+
\beta_{\mathrm{MPC}}\mathcal L_{\mathrm{MPC}}
-
\lambda_{\mathrm{ent}}\mathcal H(\pi_\theta).
\label{eq:actor_objective}
\end{equation}

The critic is updated throughout training using simulator trajectories,
separately from the world-model likelihood loss. Specifically, let $\delta_t=r_t+\gamma(1-d_t)V_{\psi_{\rm old}}(s_{t+1})
-V_{\psi_{\rm old}}(s_t)$ and compute GAE advantages $\hat A_t$ with
episode-boundary masking. The detached value target is
$\hat R_t=\hat A_t+V_{\psi_{\rm old}}(s_t)$, giving
\begin{equation}
\mathcal L_V(\psi)=\mathbb E_{t\sim\mathcal D_{\rm on}}
\left[\left(v_\psi(s_t,h_t^-)
-\operatorname{stopgrad}\left(\frac{\hat R_t-\mu_R}{\sigma_R}\right)\right)^2\right].
\label{eq:critic_loss}
\end{equation}
Here $v_\psi$ is the normalized recurrent value head, while all GAE
and planning expressions use $V_\psi=\sigma_R v_\psi+\mu_R$ in raw reward
units; memory arguments are suppressed elsewhere. Neither imagined trajectories nor historical world-model replay supply
critic targets, and the critic is held fixed during each CEM query.
We update the critic for ten epochs per collected rollout, with value
clipping at $0.2$ and running return normalization. Time-limit truncations
bootstrap from the final pre-reset observation, whereas goals and
accidental terminations disable bootstrapping; GAE traces stop at every
episode boundary.

At the beginning of training, the policy uses standard MAPPO self-play
while collecting transitions for $\mathcal D_{\mathrm{WM}}$. After a
warm-up period of $T_{\mathrm{warm}}$ environment steps, MPC guidance is
enabled, optionally with a gradual increase in its coefficient:
\begin{equation}
\beta_{\mathrm{MPC}}(n)
=
\beta_{\max}g(n),
\end{equation}
where $g(n)$ increases from zero to one after warm-up. The complete training procedure is summarized in Algorithm~\ref{alg:training}.


\begin{algorithm}[t]
\caption{Online world-model-boosted self-play training}
\label{alg:training}
\begin{algorithmic}[1]
\State Initialize decentralized actors $\pi_{\theta}$ and centralized critic
$V_{\psi}$
\State Initialize world-model ensemble $\{f_{\phi}^m\}_{m=1}^{M}$
, replay buffer $\mathcal{D}_{\mathrm{WM}}$ and opponent population $\mathcal{P}$
\For{each training interval}
    \State Sample a frozen opponent $\pi_{\mathrm{opp}}$ from $\mathcal{P}$
    \State Collect on-policy self-play trajectories using $\pi_{\theta}$ and store the transitions in $\mathcal{D}_{\mathrm{WM}}$
    
    \If{in warm-up period}
        \State Update $\pi_{\theta}$ using the MAPPO actor objective
    \Else
        \State Select training states from the collected trajectories
        \State Run hybrid CEM-MPC from each selected state
        \State Compute $\mathcal{L}_{\mathrm{MPC}}$ and update $\pi_{\theta}$ using Eq.~(\ref{eq:actor_objective})
    \EndIf
    \State Update $V_\psi$ on collected simulator trajectories using Eq.~(\ref{eq:critic_loss})
    \State Update the world-model ensemble using minibatches from $\mathcal{D}_{\mathrm{WM}}$ and $\mathcal L_{\mathrm{WM}}$
    \State Add the current checkpoint to $\mathcal{P}$ according to the population schedule
\EndFor
\end{algorithmic}
\end{algorithm}

\section{Experiments}
\label{sec:experiments}
The experiments are conducted in Isaac Gym \citep{makoviychuk2021isaac} using the \href{https://www.unitree.com/mobile/As2}{Unitree As2} robot model. Each team contains $N=2$ robots. The low-level
repositioning, dribbling, and kicking controllers are pretrained
before high-level training and remain frozen throughout all
experiments. The high-level policy operates at a fixed decision
interval, and each selected skill is executed for the corresponding
number of low-level simulation steps. Although goals are delayed,
training uses the dense shaping terms in Appendix~\ref{apdx: reward setting};
this is not a sparse-reward-only evaluation. The complete configuration and evaluation protocol are reported in
Appendices~\ref{apdx:protocol}--\ref{apdx:configuration}.

We evaluate four aspects of the framework:
1) environment-sample efficiency and head-to-head performance;
2) prediction quality and computational cost across rollout horizons;
3) the effects of online world-model updates and historical replay;
and 4) the contributions of terminal-value guidance and uncertainty
penalization.


\subsection{Main Results}
Appendix~\ref{apdx:qualitative} illustrates a representative attacking sequence.

\begin{figure}[t!]
\centering
\subfigure[]{\includegraphics[width=0.45\linewidth]{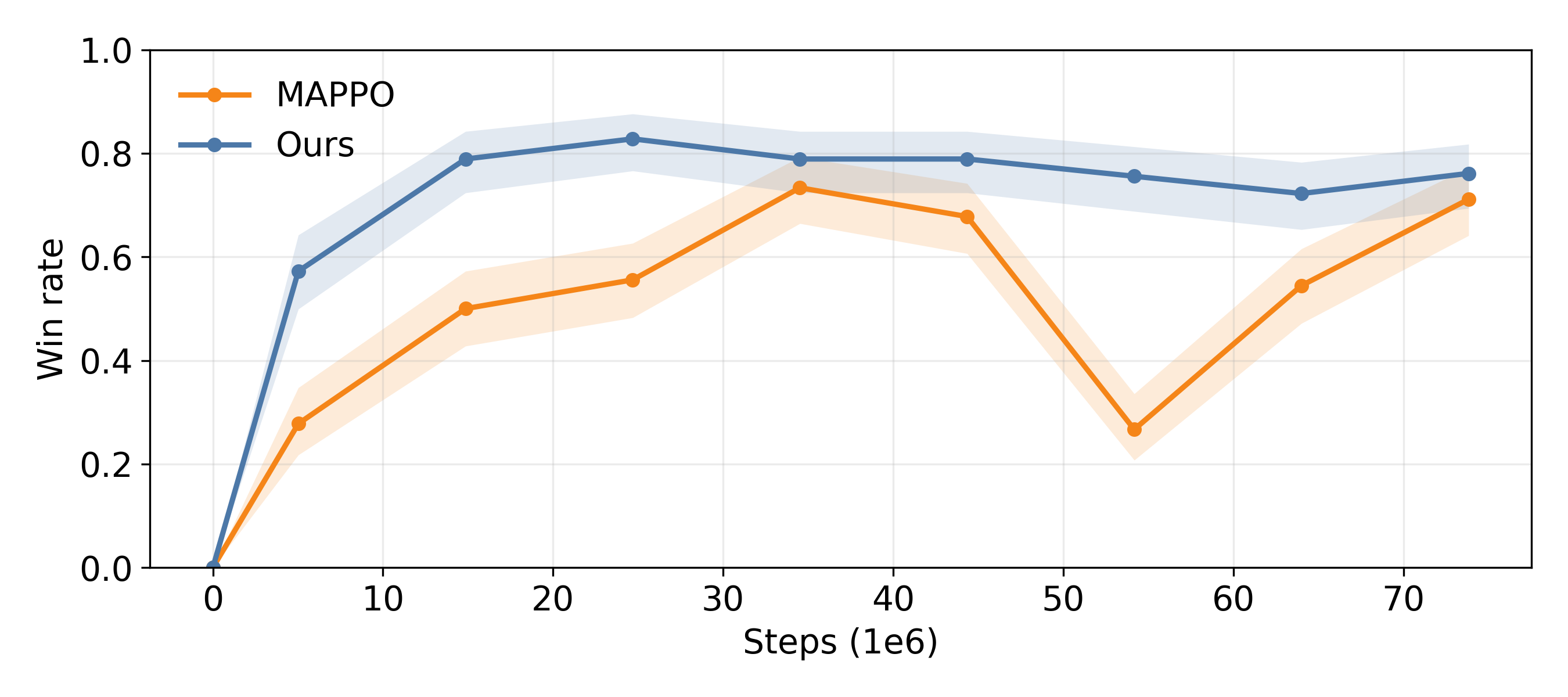}}
\subfigure[]{\includegraphics[width=0.45\linewidth]{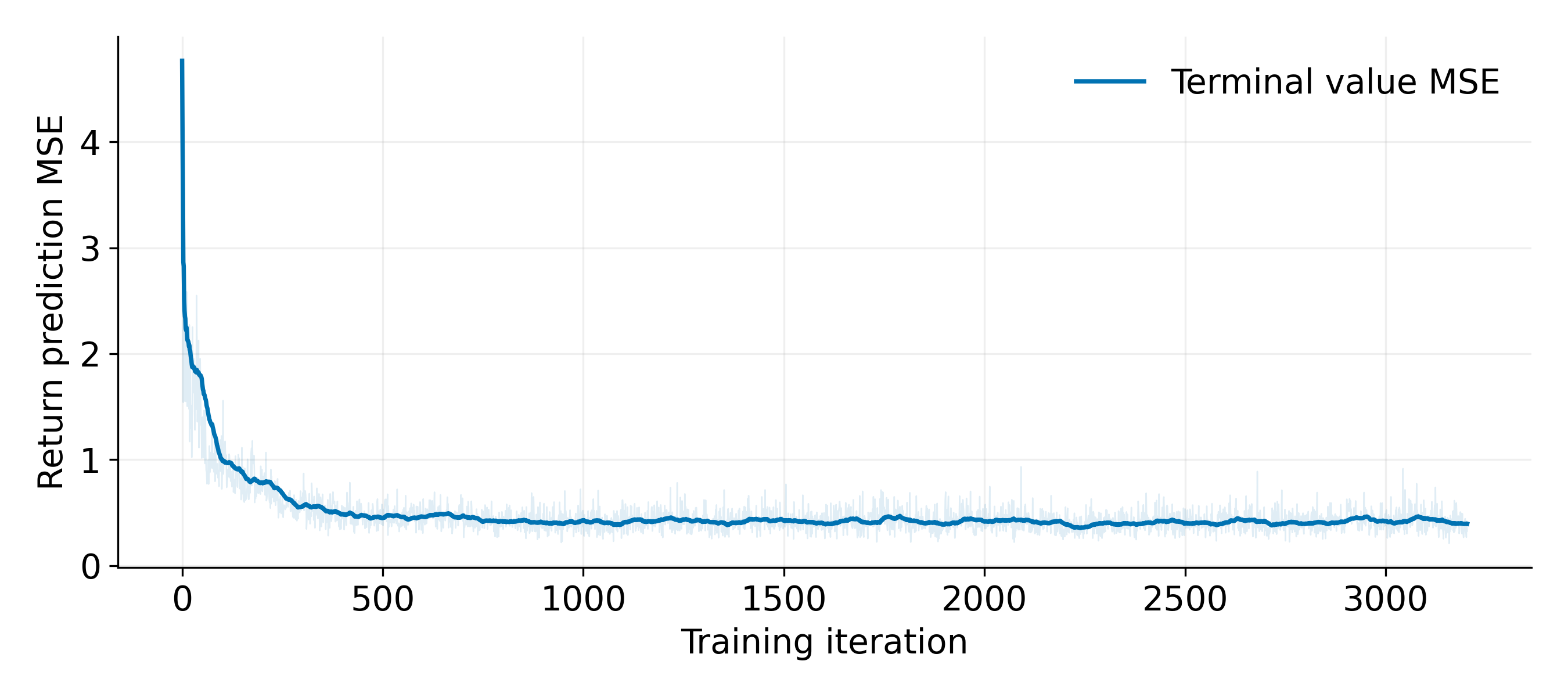}}
\subfigure[]{\includegraphics[width=0.45\linewidth]{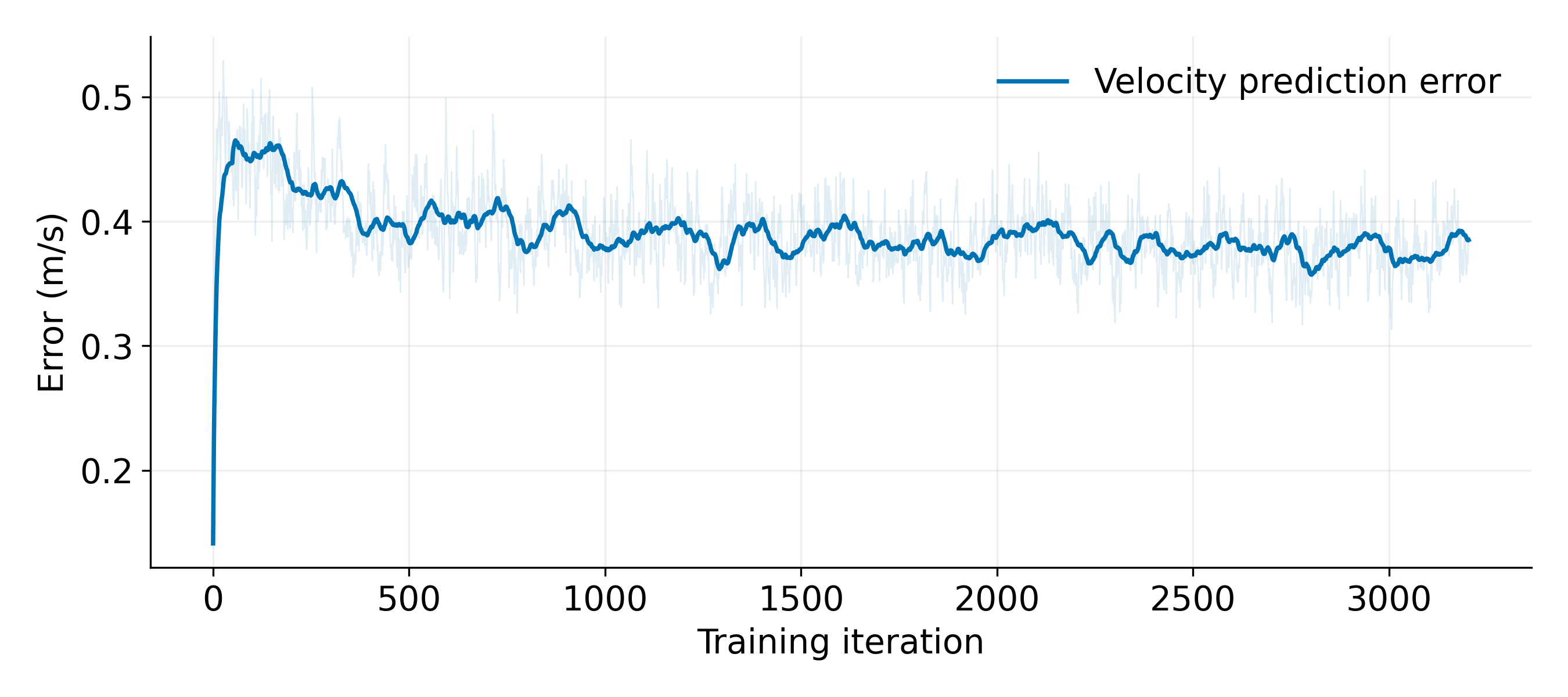}}
\subfigure[]{\includegraphics[width=0.45\linewidth]{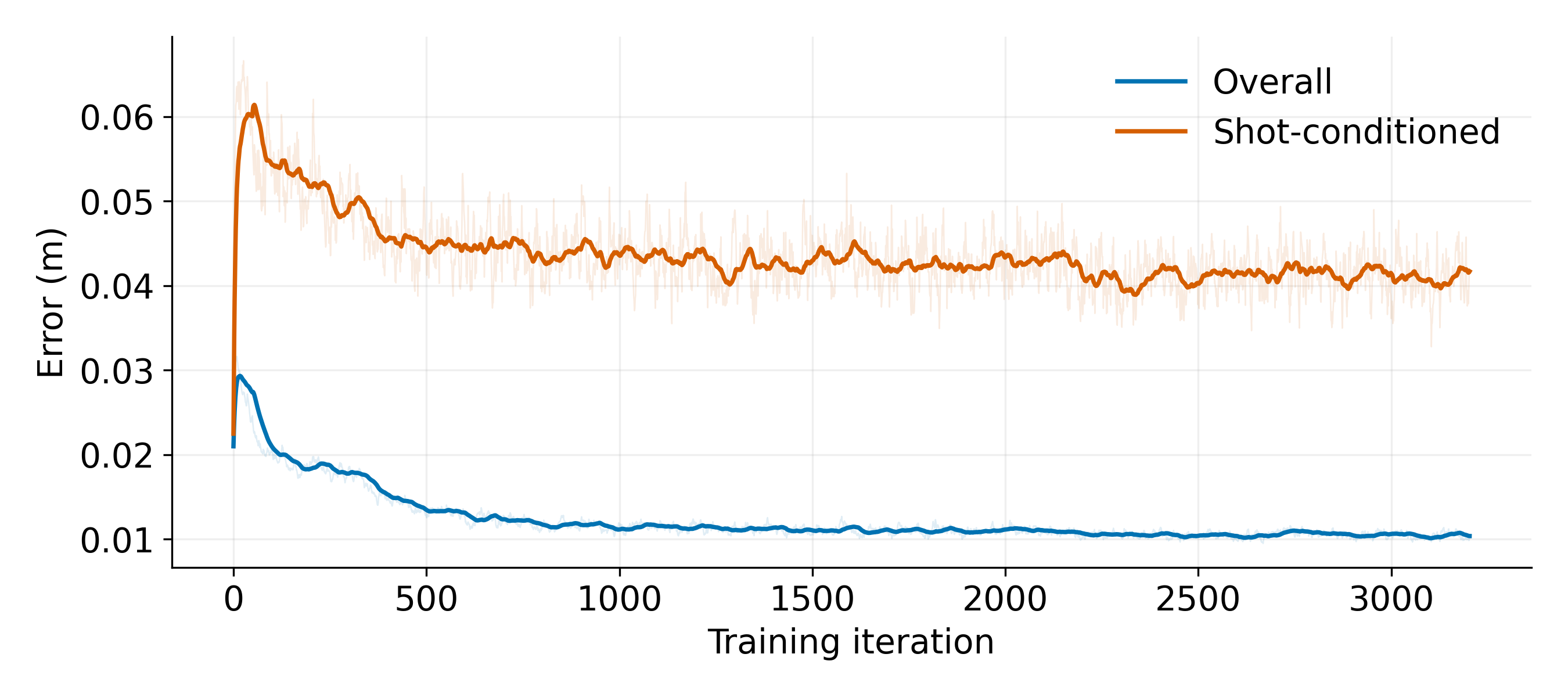}}
\caption{Training dynamics (comparison run C75, up to 75M joint transitions). (a) Fixed-pool all-match win rate; this run differs from the 40M ablation run in Figure~\ref{fig:ablation}. (b) Terminal-value prediction error. (c) Velocity prediction error. (d) Ball-position prediction error, reported for all transitions and for shot-conditioned transitions separately.}
\label{fig:learning}
\end{figure}

Figure~\ref{fig:learning} (a) compares win-rate learning curves under the same fixed opponent pool. Our method shows faster early learning and a higher final win rate than MAPPO under the same fixed evaluation pool. The narrowing gap later in training suggests that additional model-free interaction partially closes the difference, whereas the early improvement is consistent with the benefit of world-model-boosted policy learning. Figures~\ref{fig:learning} (b)-(d) show the corresponding prediction errors. As the policies learn basic coordination, the self-play data become more strongly coupled and harder to predict. Online updates then reduce the position-prediction error, while the velocity error remains higher because kicking causes abrupt ball-motion changes.

We further compare the methods in head-to-head matches, as reported in Table~\ref{tab:compete}. For each matchup, we conduct 100 matches per seed across 10 random seeds and report the mean decisive-match conditional win rate,
$100W_s/(W_s+L_s)$, across seeds $s$; draws are excluded from this
denominator.  Appendix~\ref{apdx:statistics}
reports an additional independent evaluation with per-seed W/D/L counts
and both all-match and decisive-match statistics. In these matches, our method wins against MAPPO and discrete MAPPO with rates of $59.47\%$ and $93.72\%$, respectively, while MAPPO wins against discrete MAPPO at $92.43\%$. These results indicate a moderate advantage over continuous-action MAPPO and a much larger gap between the continuous- and discrete-action formulations.

Discrete MAPPO uses eight fixed-direction commands with a fixed velocity magnitude. Although this removes the continuous command variables, it restricts the available direction and speed choices, which can limit precise coordination. We implemented this baseline based on \citet{pmlrv305su25a}; its results may therefore differ from the original work.

\begin{table}[htp]
\centering
\caption{Decisive-match conditional win rate against different opponents}
\label{tab:compete}
\small
\begin{tabular}{cccc}
\toprule
\multicolumn{3}{c}{Evaluated match} & Win rate \\
\midrule
Ours & v.s. & MAPPO & \textbf{$(59.47\pm0.92)\%$} \\
Ours & v.s. & Discrete MAPPO & $(93.72\pm0.54)\%$ \\
MAPPO & v.s. & Discrete MAPPO & $(92.43\pm0.56\%)$ \\
\bottomrule
\end{tabular}
\end{table}

\subsection{Planning Horizon and World-Model Quality}
Short model rollouts balance prediction accuracy and planning cost.
The default $H=4$ achieves the highest policy win rate under both matched
interactions and matched wall time (Appendix~\ref{apdx:model_diagnostics},
Table~\ref{tab:model} and ~\ref{tab:horizon_policy}). Historical replay
reduces position error from $14.0$ cm (Frozen) and $18.1$ cm (Online-only)
to $11.6$ cm. In the A40 comparisons, final all-match win rates are
$68.0\%$, $64.0\%$, and $80.0\%$, respectively. The stratified diagnostics connect this benefit to
retention of historical interactions and improved candidate ranking,
rather than prediction error alone.

\subsection{Ablation Study}

Table~\ref{tab:main_ablation} isolates the main components using the same
continuous actions, evaluation pool, and interaction budget. Removing
terminal value, uncertainty regularization, joint planning, or historical
replay lowers final performance. 
Joint- versus independent-teacher students provide the direct comparison of planning and coordination transfer.

\input{generated/main_ablation.tex}

\begin{figure}[htp]
    \centering
    \subfigure[]{\includegraphics[width=0.45\linewidth]{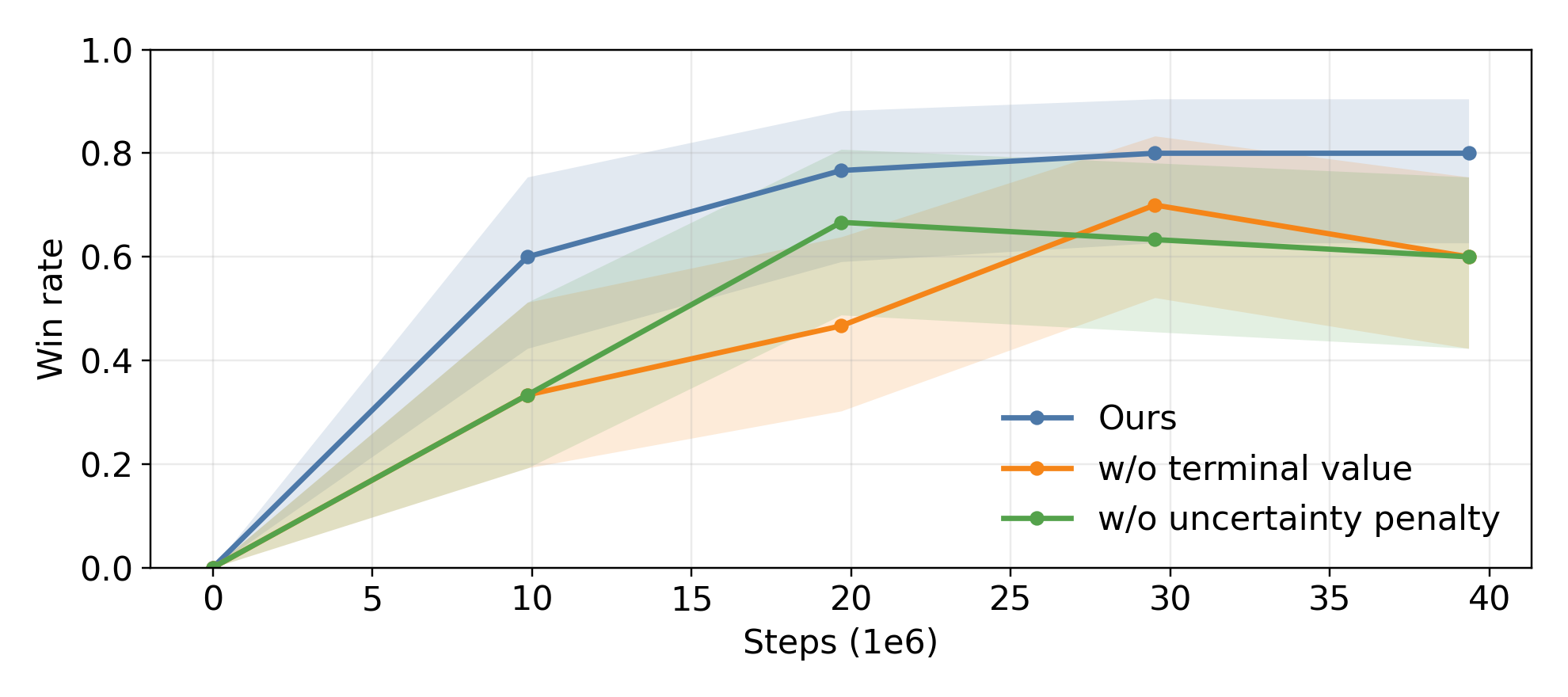}}
    \subfigure[]{\includegraphics[width=0.45\linewidth]{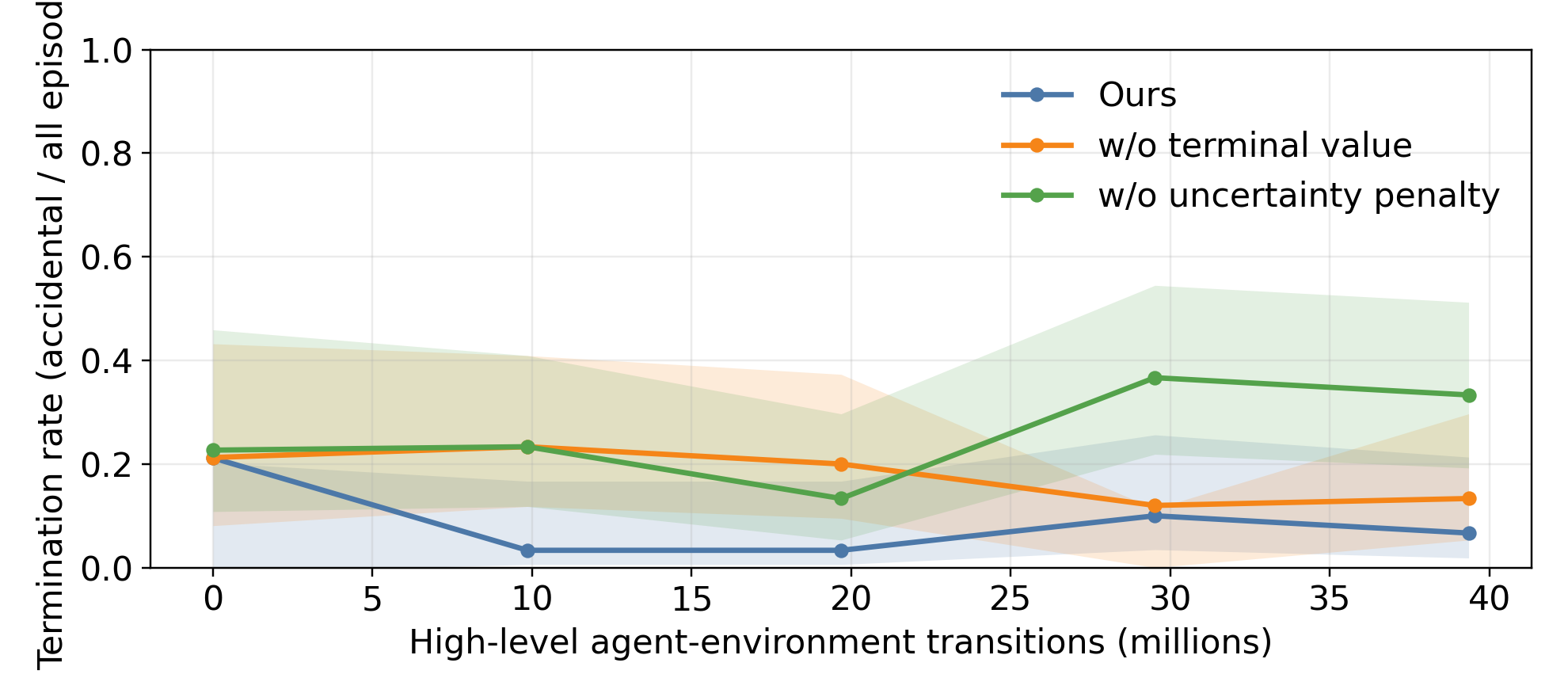}}
    \caption{Ablation of terminal-value guidance and uncertainty
    penalization (run A40, 40M joint transitions). (a) Fixed-pool all-match win rate. (b) Accidental-termination rate, defined as the fraction of evaluation episodes ending in an accidental termination. }
    \label{fig:ablation}
\end{figure}

Figure~\ref{fig:ablation} compares the learning curves and termination
rates of the three variants. Removing the terminal-value term leads to
slower early improvement and lower final win rate. This result is
consistent with the role of the terminal value in providing an estimate
of the return beyond the finite model rollout, thereby reducing the
myopic bias of short-horizon reward optimization. In contrast, the full
method reaches a higher win rate earlier and maintains the best
performance throughout the later stages of training.

Removing the uncertainty penalty produces a larger termination rate as shown in Figure~\ref{fig:ablation},
particularly after the policy begins to acquire more effective
interaction behaviors. This suggests that, without uncertainty
regularization, the MPC teacher is more likely to select candidate
sequences that rely on poorly modeled state-action regions. The
uncertainty penalty suppresses such model exploitation and results in
more stable training. The full method combines both effects: the
terminal value provides long-term guidance for short-horizon planning,
while the uncertainty penalty improves the reliability of model-based
action evaluation.

\section{Conclusion}
We presented an online world-model-boosted framework for multi-agent
quadruped soccer with hybrid discrete-continuous actions.
A probabilistic ensemble world model is continually updated using
self-play transitions and historical replay to capture interactions
among robots and the ball. An uncertainty-aware hybrid MPC teacher
combines short model rollouts, terminal-value guidance, and
frozen-opponent policy predictions to optimize skill selections
and continuous commands. Its action distributions provide auxiliary
KL supervision for MAPPO actors, while environment actions remain
policy-generated. Deployment requires only the decentralized actors
and frozen low-level skills.

Experiments show faster early learning and an advantage over MAPPO. World-model prediction comparisons favor combining online updates with historical replay, while ablations show that both terminal-value guidance and uncertainty penalization contribute to higher final win rates within the evaluated environment-interaction budget. Additional controls demonstrate improved fixed-pool and unseen-opponent results at matched interactions, with metric-specific benefits of joint-teacher students.  
\label{sec:main_end}

\section{REPRODUCIBILITY STATEMENT}
We are committed to ensuring the reproducibility of all results presented in this paper. To facilitate reproduction, we publicly release all source code, including those to train the high-level policy, low-level skills, and world model, at our project page:\\
\vspace{-0.6cm}
\begin{center}
\url{https://anonymous.4open.science/w/skills2strategy}.
\end{center}


\section{LLM Usage}
LLMs were not used in the conceptualization of the study, formulation of its key contributions, or identification of relevant prior work. They were used for assistance with code debugging and manuscript organization and revision. The authors reviewed all suggestions and take responsibility for the scientific claims, experimental results, and final manuscript.

\bibliography{iclr2027_conference}
\bibliographystyle{iclr2027_conference}
\newpage
\appendix
\section{Appendix reward setting}
\label{apdx: reward setting}
\begin{table}[h]
  \centering
  \small
  \caption{Low level reward setting}
  \label{tab: low level reward}
  \label{tab:skill_rewards}
  \begin{tabularx}{\textwidth}{l l X r}
  \toprule
  \textbf{Skill} & \textbf{Reward term}
  & \textbf{Brief explanation} & \textbf{Coefficient} \\
  \midrule

  Common & \texttt{orientation}
  & Penalize body roll and pitch. & $-5.0$ \\
  & \texttt{torques}
  & Penalize squared joint torques. & $-3\times10^{-5}$ \\
  & \texttt{dof\_vel}
  & Penalize squared joint velocities. & $-10^{-4}$ \\
  & \texttt{dof\_acc}
  & Penalize squared joint accelerations. & $-2.5\times10^{-7}$ \\
  & \texttt{collision}
  & Penalize contacts on selected robot bodies. & $-1.0$ \\
  & \texttt{action\_rate}
  & Penalize changes between consecutive actions. & $-0.01$ \\
  & \texttt{contacts\_shaped\_force}
  & Penalize foot contact forces during swing. & $4.0$ \\
  & \texttt{contacts\_shaped\_vel}
  & Penalize foot motion during stance. & $4.0$ \\
  & \texttt{dof\_pos\_limits}
  & Penalize joint position limit violations. & $-10.0$ \\
  & \texttt{dof\_pos}
  & Penalize deviations from default joint positions. & $-0.05$ \\
  & \texttt{action\_smoothness\_1}
  & Penalize first differences of joint position targets. & $-0.1$ \\
  & \texttt{action\_smoothness\_2}
  & Penalize second differences of joint position targets. & $-0.1$ \\

  \midrule
  Reposition & \texttt{walking\_vel}
  & Track commanded planar velocity and yaw rate. & $4.0$ \\

  \midrule
  Dribble & \texttt{robot\_ball\_vel}
  & Encourage approaching the ball at the commanded speed. & $0.5$ \\
  & \texttt{robot\_ball\_pos}
  & Keep the ball near the desired position ahead of the robot. & $4.0$ \\
  & \texttt{backward\_motion}
  & Penalize robot motion opposite the requested direction. & $2.0$ \\
  & \texttt{ball\_vel}
  & Track the commanded planar ball velocity. & $4.0$ \\
  & \texttt{robot\_ball\_yaw}
  & Keep the ball aligned with the robot's forward direction. & $4.0$ \\
  & \texttt{ball\_vel\_norm}
  & Track the commanded ball speed. & $4.0$ \\
  & \texttt{ball\_vel\_angle}
  & Align ball motion with the commanded direction. & $4.0$ \\
  & \texttt{tracking\_ang\_vel}
  & Track the commanded robot yaw rate. & $1.0$ \\

  \midrule
  Kick & \texttt{ball\_vel}
  & Track target ball velocity after the kick and separation. & $5.0$ \\
  & \texttt{ball\_forward}
  & Reward ball motion along the commanded shot direction. & $8.0$ \\
  & \texttt{ball\_vel\_norm}
  & Track target ball speed after the kick and separation. & $1.5$ \\
  & \texttt{ball\_vel\_angle}
  & Align post-kick ball motion with the shot direction. & $1.5$ \\
  & \texttt{ball\_out}
  & Reward ball travel and speed along the shot direction. & $7.0$ \\
  & \texttt{robot\_approach\_ball}
  & Reward signed progress toward the pre-kick setup position. & $1.5$ \\
  & \texttt{excess\_yaw}
  & Penalize excessive yaw rate before the kick. & $-0.1$ \\
  & \texttt{launch}$^{*}$
  & Reward a qualifying ball launch. & $5.0$ \\
  & \texttt{success}$^{*}$
  & Reward meeting shot speed, alignment, and travel criteria. & $25.0$ \\
  & \texttt{failure}$^{*}$
  & Penalize a failed shooting attempt. & $-2.0$ \\

  \bottomrule
  \end{tabularx}
  *Starred terms are event rewards; other terms are timestep-scaled.
  \end{table}

\begin{table}[h]
  \centering
  \caption{High level reward setting}
  \label{tab:high_level_rewards}
  \begin{tabularx}{\textwidth}{l X r}
  \toprule
  \textbf{Reward term} & \textbf{Brief explanation} & \textbf{Coefficient} \\
  \midrule
  \texttt{goal}$^{*}$
  & Score in the opponent's goal. & $75.0$ \\
  \texttt{opponent\_goal}$^{*}$
  & Concede a goal. & $-75.0$ \\
  \texttt{accidental\_termination}$^{*}$
  & Learning-team failure causing termination. & $-5.0$ \\
  \texttt{ball\_goal\_progress}
  & Signed ball progress toward the goal. & $2.0$ \\
  \texttt{goalward\_ball\_velocity}
  & Ball velocity directed toward the goal. & $0.75$ \\
  \texttt{approach\_ball}
  & Attacker approaches the ball from outside skill range. & $1.0$ \\
  \texttt{dribble\_ball\_control}
  & Controlled, goalward dribbling. & $1.0$ \\
  \texttt{shoot\_launch}$^{*}$
  & Valid shot launches the ball toward the goal. & $5.0$ \\
  \texttt{invalid\_skill}
  & Penalize invalid skill requests. & $-2.0$ \\
  \texttt{robot\_collision}
  & Penalize unsafe clearance between robots. & $-2.0$ \\
  \texttt{support\_lane}
  & Penalize support robots obstructing the shot lane. & $-0.5$ \\
  \texttt{attack\_position}
  & Attacker progresses toward a position behind a slow ball. & $1.0$ \\
  \bottomrule
  \end{tabularx}
  *Starred terms are event rewards; other terms are timestep-scaled.
  \label{tab: high level reward}
  \end{table}
\FloatBarrier
\section{Continuous Command Setting}
\label{apdx:low level skills}
\begin{table}[h]
  \centering
  \caption{Low-level skills and their continuous command spaces.}
  \label{tab:low_level_skills}
  \begin{tabular}{lll}
  \toprule
  Low-level skill & Continuous output definition & Range \\
  \midrule
  Reposition &
  \makecell[l]{Robot planar velocity $(v_x, v_y)$\\
  Robot yaw rate $\omega_z$} &
  \makecell[l]{$v_x,v_y \in [-1.5,1.5]\,\mathrm{m/s}$\\
  $\omega_z \in [-1,1]\,\mathrm{rad/s}$} \\
  \addlinespace
  Dribble &
  \makecell[l]{Target ball velocity $(v_x^{b},v_y^{b})$\\
  Robot yaw rate $\omega_z$} &
  \makecell[l]{$v_x^{b},v_y^{b} \in [-1.5,1.5]\,\mathrm{m/s}$\\
  $\omega_z \in [-1,1]\,\mathrm{rad/s}$} \\
  \addlinespace
  Kick &
  Target ball velocity $(v_x^{b},v_y^{b},0)$ &
  $v_x^{b},v_y^{b} \in [-3,3]\,\mathrm{m/s}$ \\
  \bottomrule
  \end{tabular}\\
  *The third command component is unused for kicking.
  \end{table}
\input{rebuttal_additions.tex}

\end{document}

%% file: math_commands.tex
\usepackage{amsmath,amsfonts,bm}

\def\eqref#1{equation~\ref{#1}}

\def\1{\bm{1}}

\DeclareMathAlphabet{\mathsfit}{\encodingdefault}{\sfdefault}{m}{sl}
\SetMathAlphabet{\mathsfit}{bold}{\encodingdefault}{\sfdefault}{bx}{n}



%% file: generated/main_ablation.tex
\begin{table}[htbp]
\centering\small
\setlength{\tabcolsep}{4pt}
\caption{Ablation of the main components in A40. All-match win rate (\%); mean $\pm$ sample SD over ten seeds. $B=40$M joint transitions.}
\label{tab:main_ablation}
\begin{tabular}{lccc}
\toprule
Variant & 10M & 20M & 40M \\
\midrule
Full method & $60.0\pm15.7$ & $76.7\pm13.6$ & $80.0\pm10.5$ \\
Without terminal value & $33.3\pm18.9$ & $46.7\pm16.8$ & $60.0\pm16.8$ \\
Without uncertainty penalty & $33.3\pm17.8$ & $66.7\pm13.6$ & $60.0\pm17.8$ \\
Without distillation & $40.0\pm12.6$ & $55.0\pm11.5$ & $70.0\pm11.5$ \\
Independent MPC teacher & $52.0\pm12.6$ & $65.0\pm11.5$ & $72.0\pm11.5$ \\
Online-only world model & $44.0\pm12.6$ & $57.0\pm11.5$ & $64.0\pm11.5$ \\
Frozen world model & $48.0\pm12.6$ & $60.0\pm11.5$ & $68.0\pm11.5$ \\
Without teacher mask & $57.0\pm12.6$ & $72.0\pm11.5$ & $77.0\pm11.5$ \\
\bottomrule
\end{tabular}
\end{table}

%% file: rebuttal_additions.tex
\FloatBarrier
\section{Targeted Controls and Evaluation Details}
\label{apdx:controls}
\subsection{Attributing the gains}
\label{apdx:protocol}
All added controlled comparisons use the A40 protocol: $B=40$ million
joint high-level environment transitions, training seeds 0--9, the same
fixed evaluation pool, and all-match win rate $W/(W+D+L)$. One joint
transition advances all four robots once; it is not multiplied by the
number of robots or low-level controller steps. 

\paragraph{Controls}
All methods share skills, observations, rewards, recurrent architectures,
training seeds, and the evaluation pool. MAPPO + feasibility prior uses
the same distance mask for actor sampling and PPO likelihoods; the
no-mask ablation removes it only from the teacher. The one-step teacher
uses $H=1$ and reallocates the $H=4$ model-evaluation budget to more
candidates. Reward-only MPC removes terminal value and uncertainty.
Independent MPC fixes teammate actions to actor predictions and splits
the joint teacher's model-evaluation budget equally between two searches.
The same global information is available to both teachers. All policy
results evaluate decentralized students, not online MPC controllers.
Frozen fits on a common warm-up dataset; Online-only uses recent data;
replay mixes recent and historical transitions with matched update counts.

\input{generated/learning_controls.tex}

At 40M, the joint-teacher student reaches $80.0\%$, versus $72.0\%$ for
the independent-teacher student, $72.0\%$ for MAPPO with the prior,
$73.0\%$ for the one-step teacher, and $55.0\%$ for reward-only planning.
The full method exceeds Frozen ($68.0\%$) and Online-only ($64.0\%$).
These are descriptive differences with substantial across-seed variation,
not a claim of significance from mean gaps alone. Setting the KL weight
to zero disconnects the only model-to-actor channel; returning to MAPPO
therefore checks this pathway, not a novel distillation mechanism.

\paragraph{Candidate ranking and coordination.}
We evaluated 1,000 held-out states per seed and 256 shared candidates
per state with matched opponent randomness, horizon, and terminal critic.
Selection regret is $(J_{\max}^{\rm sim}-J_{\rm selected}^{\rm sim})/
(J_{\max}^{\rm sim}-J_{\min}^{\rm sim}+10^{-8})$. Half the states are
recent and half historical. Recent/historical position errors are
$13.0/15.0$ cm (Frozen), $10.2/26.0$ cm (Online-only), and $10.4/12.8$ cm
(Replay), whose means match the $14.0,18.1,11.6$ cm results.

Support-lane obstruction measures valid kicks with a teammate in the
$0.5$ m-wide ball-to-goal corridor. Recovery measures blocked shots
followed by teammate possession within 2 s; possession requires ball
distance below $0.5$ m and relative speed below $0.5$ m/s for three
decisions. Each seed record contains 1,000 eligible events per
metric. Joint- and independent-teacher students share evaluation seeds.
\begin{table}[htbp]
\centering\small
\caption{Candidate ranking on a shared held-out candidate set; mean
$\pm$ standard deviation over ten seeds. These metrics evaluate the
world model used by a teacher.}
\label{tab:ranking_controls}
\begin{tabular}{lcc}
\toprule
World model & Ranking $\rho\uparrow$ & Normalized regret $\downarrow$\\
\midrule
Frozen & $0.61\pm0.04$ & $0.19\pm0.03$\\
Online-only & $0.54\pm0.05$ & $0.24\pm0.04$\\
Online + Historical Replay & $0.82\pm0.03$ & $0.08\pm0.02$\\
\bottomrule
\end{tabular}
\end{table}
\input{generated/coordination.tex}

The joint-teacher student improves obstruction from $13.0\%$ to $7.0\%$
and recovery from $49.0\%$ to $61.0\%$ relative to the independent-teacher
student. Relative to MAPPO, the obstruction improvement is
$(18-7)/(18-6)=91.7\%$; the recovery improvement is
$(61-42)/(64-42)=86.4\%$. These are ratios of mean improvements for
two specific metrics, not a fraction of overall coordination ability.
Paired seed bootstrap intervals (10,000 resamples) computed from the
event records are $[87.6,96.2]\%$ and $[81.7,91.5]\%$,
respectively.

\subsection{Evaluation statistics and opponent separation}
\label{apdx:statistics}
The results use seed means of $100W/(W+D+L)$ with sample SD ($n-1$
denominator), not standard error. Decisive-match rates use $100W/(W+L)$.
\begin{table}[htbp]
\centering\small
\caption{Per-seed W/D/L in the additional independent evaluation.}
\label{tab:seed_counts}
\begin{tabular}{rccc}
\toprule
Seed & Ours vs. MAPPO & Ours vs. discrete & MAPPO vs. discrete\\
\midrule
1 & 50/14/36 & 88/6/6 & 86/7/7\\
2 & 52/13/35 & 89/5/6 & 87/6/7\\
3 & 49/16/35 & 87/7/6 & 85/8/7\\
4 & 54/12/34 & 90/4/6 & 88/5/7\\
5 & 51/15/34 & 88/6/6 & 86/7/7\\
6 & 50/14/36 & 89/5/6 & 87/6/7\\
7 & 53/13/34 & 87/7/6 & 85/8/7\\
8 & 48/17/35 & 90/4/6 & 88/5/7\\
9 & 52/14/34 & 88/6/6 & 86/7/7\\
10 & 51/15/34 & 89/5/6 & 87/6/7\\
\midrule
Total & 510/143/347 & 885/55/60 & 865/65/70\\
\bottomrule
\end{tabular}
\end{table}

Table~\ref{tab:seed_counts} is a separate 100-match-per-seed
head-to-head evaluation: totals are $510/143/347$, $885/55/60$, and
$865/65/70$; mean decisive win rates are $59.50\pm1.28\%$,
$93.65\pm0.07\%$, and $92.51\pm0.09\%$. These are not the raw counts
underlying Table~\ref{tab:compete}.

\paragraph{Pools and seed records.}
The fixed pool contains checkpoints at 8, 16, 24, 32, and 40M from
independent MAPPO seeds 100 and 101. Unseen opponents use seeds 200 and
201 and are excluded from training, replay, and selection. New fixed-
and unseen-pool evaluations use 1,000 matches per training seed: 100
per checkpoint, equally split between field sides. The old head-to-head
protocol uses 100 matches and is a different evaluation. Matches last
at most 60 s and stop at the first goal or accidental termination;
timeouts without goals and simultaneous failures are draws. Individual failures count as losses for the failing team. Final checkpoints are used unless a matched-time cutoff is explicitly stated.

Evaluation seeds are paired across methods at the same checkpoint and pool. Table~\ref{tab:pool_seed_counts} gives the central 40M counts.

\FloatBarrier
\subsection{Configuration and total computation}
\label{apdx:configuration}
\begin{table}[htbp]
\centering\small
\caption{Training and planning configuration shared across controls
except for explicitly ablated components.}
\begin{tabularx}{\linewidth}{lX}
\toprule
Setting & Value\\
\midrule
Skill / low-level interval & $0.1$ s / $0.02$ s; five controller steps per decision\\
World model & Five independent MLPs; three 256-unit SiLU layers; normalized state inputs including boundary joint positions/velocities; residual, raw reward, termination heads\\
Model optimizer / loss & Adam $10^{-3}$; batch 1024; $\lambda_r=\lambda_d=1$; log-variance bounded to $[-10,2]$; independent bootstrap minibatches\\
Model update / replay & 50 minibatches per rollout; capacity $10^6$ joint transitions; 50:50 latest rollout / older replay\\
Actor / critic & Separate two-layer 128-unit MLPs and 128-unit GRUs; parameter sharing within each team; running input normalization; normalized value head de-normalized before planning\\
PPO & Adam $3\times10^{-4}$; $\gamma=0.99$, GAE $\lambda=0.95$, policy/value clip $0.2$, entropy weight $0.01$; ten epochs, four minibatches; gradient norm $0.5$\\
Collection / budget & 256 environments; 128 high-level steps per rollout; $B=4\times10^7$ joint transitions (A40), without multiplication by robot count\\
Warm-up / distillation & $T_{\rm warm}=10^6$; $\beta_{\max}=0.1$, $\lambda_c=1$; linear ramp over the next $10^6$ transitions\\
Teacher queries & 256 on-policy states per rollout after warm-up; states processed sequentially; 256 candidates and five members vectorized within a state; cached targets\\
CEM & $H=4$; 256 candidates; 32 elites; four iterations; all five model members; new elite moments weighted by $\alpha=0.1$\\
CEM distributions & Actor initialization; commands clipped to skill bounds; normalized std floor $0.05$; feasible-skill probability floor $10^{-3}$; retain moments for skills without elites\\
Uncertainty / feasibility & $\lambda_u=0.1$; $W_s$ is diagonal inverse state variance; dribble/kick allowed within $0.5$ m of the ball; reposition always allowed\\
Opponent population & Uniform sampling; one checkpoint per $10^6$ transitions; 100-checkpoint FIFO capacity; initialized with the random high-level policy\\
Hardware / software & One NVIDIA RTX 4090 (24 GB), AMD Ryzen 9 7950X, 64 GB RAM; Ubuntu 22.04, CUDA 11.8, PyTorch 2.0.1, Isaac Gym Preview 4\\
\bottomrule
\end{tabularx}
\end{table}

For the Su et al. adaptation, we retain frozen skills, parameter-shared
recurrent MAPPO, and historical self-play, with common simulator,
observation, reward, and checkpoint settings. Its eight-direction
commands use $1$ m/s for reposition/dribble and $3$ m/s for kick, with
yaw tracking the selected heading.

\paragraph{Auditable cost accounting.}
The latency in Table~\ref{tab:model} is 41.9 ms for one
four-step evaluation of 256 candidates from one state over five members.
Candidates and ensemble members are vectorized on one GPU; selected
states are processed sequentially, and CEM rounds are sequential.
The timing includes opponent observations/actions, dynamics, recurrent
critic propagation, and terminal scoring, but excludes CEM refitting,
query preparation, model training, and PPO. Calls are synchronized;
no multi-state speedup or early-termination throughput saving is assumed.
Terminated branches are masked, while dense batches retain their shape.

With 32,768 joint transitions per collection interval, 40M requires
1,221 intervals (the last is partial). The first 30 complete intervals
are below the 1M warm-up; the remaining 1,191 invoke the teacher.
Thus the rollout cost is
\[
\frac{1191\times256\times4\times0.0419}{3600}=14.195\text{ h}.
\]
Adding 0.8 h for state preparation, CEM refitting and target storage
gives 14.995 h of teacher work. Simulation (4.8 h), PPO (2.4 h), model
fitting (1.2 h), and evaluation (0.8 h) give 24.195 h total, versus
8.0 h for MAPPO. At the 100M budget, the same formula gives approximately 36 h of rollout work. Shared skill pretraining (48 GPU-hours) is separate.

\input{generated/compute.tex}

At matched 40M interactions, fixed-pool win rates are $80.0\%$ and
$70.0\%$; unseen W/D/L are $75/8/17\%$ and $65/8/27\%$ for Full and
MAPPO. At 24.2 h, extended MAPPO reaches $78.0\%$ on the fixed pool
(121M transitions), leaving only a 2-point mean gap with overlapping
seed variability. At 8 h, Full reaches $65.3\%$ (13.2M transitions),
below MAPPO's $70.0\%$. Consequently the evidence supports interaction
efficiency, but not a uniform compute-efficiency advantage. Matched-time
transition counts are separate records, not an extrapolation
of the preserved C75 curve. Memory estimates remain 19.6 GB (Full) and
13.2 GB (MAPPO).

\input{generated/horizon.tex}

The horizon sweep fixes candidate count and iteration count, whereas
the one-step control in Table~\ref{tab:additional_controls} reallocates
the full model-evaluation budget. Costs for all horizons use the same
call count and their respective latencies in
Table~\ref{tab:model}; $H=4$ gives the best mean in this sweep.

\FloatBarrier
\section{Additional Model Diagnostics and Qualitative Results}
\subsection{Planning Horizon and World-Model Quality}\label{apdx:model_diagnostics}

We next examine whether the learned world model and terminal value function provide a reliable basis for short-horizon planning. Table~\ref{tab:model} reports position-prediction error, ensemble uncertainty, terminal value MSE, and rollout time for different planning horizons. The position error contains the error for ball and robots unlike Figure \ref{fig:learning} (d). Longer rollouts incur greater prediction error and computational cost. However, optimizing only rewards within a short horizon can favor actions with immediate benefits over those whose rewards
occur later.

The terminal value term addresses this limitation by estimating the remaining return beyond the model rollout. 
Consequently, the planner can compare candidate sequences using both their short-term rewards and the estimated long-term value of their endpoint states, without requiring a very large planning horizon \citep{chen1998quasi}. The terminal value does not guarantee globally optimal planning; instead, its effectiveness depends on the accuracy of the learned critic and the world-model rollout.

\begin{table}[htp]
\centering
\caption{World-model accuracy across planning horizons.}
\label{tab:model}
\small
\setlength{\tabcolsep}{4pt}
\begin{tabular}{ccccc}
\toprule
Horizon & Position Error (cm) & Uncertainty & Terminal value MSE &
Rollout Time (ms/batch)\\
\midrule
1  & 4.1  & 1.3  & 0.71 & 10.8\\
2  & 8.7  & 2.5  & 0.89   & 21.4\\
4  & 11.6 & 4.6  & 1.16  & 41.9\\
10 & 17.6 & 15.3 & 1.68   & 105.8\\
\bottomrule
\end{tabular}
\end{table}

A key design choice is to train the world model continuously together with the evolving self-play policy. Table~\ref{tab:onlinewm} compares Frozen, Online-only, and Online + Historical Replay. The combined update has the lowest prediction error. Online-only performs worse than Frozen. The stratified evaluation in Appendix~\ref{apdx:controls} shows that recent-only updates improve recent-state prediction but degrade historical-state coverage, consistent with reduced retention of historical interactions. Historical replay improves candidate ranking and final policy performance together, linking predictive accuracy to useful planning supervision. Ensemble disagreement is treated as a planning regularizer rather than a calibrated probability of error.

\begin{table}[htp]
\centering
\caption{World model training methods comparison}
\label{tab:onlinewm}
\small
\begin{tabular}{lcc}
\toprule
World Model & Position Error (cm) &Uncertainty  \\
\midrule
Frozen                 & 14.0 & 10.8 \\
Online-only            & 18.1 & 12.9\\
\textbf{Online + Historical Replay}& \textbf{11.6} &\textbf{4.6}  \\
\bottomrule
\end{tabular}
\end{table}

\FloatBarrier
\subsection{Qualitative Attacking Sequence}
\label{apdx:qualitative}
\begin{figure}[t!]
    \centering
    \includegraphics[width=0.9\linewidth]{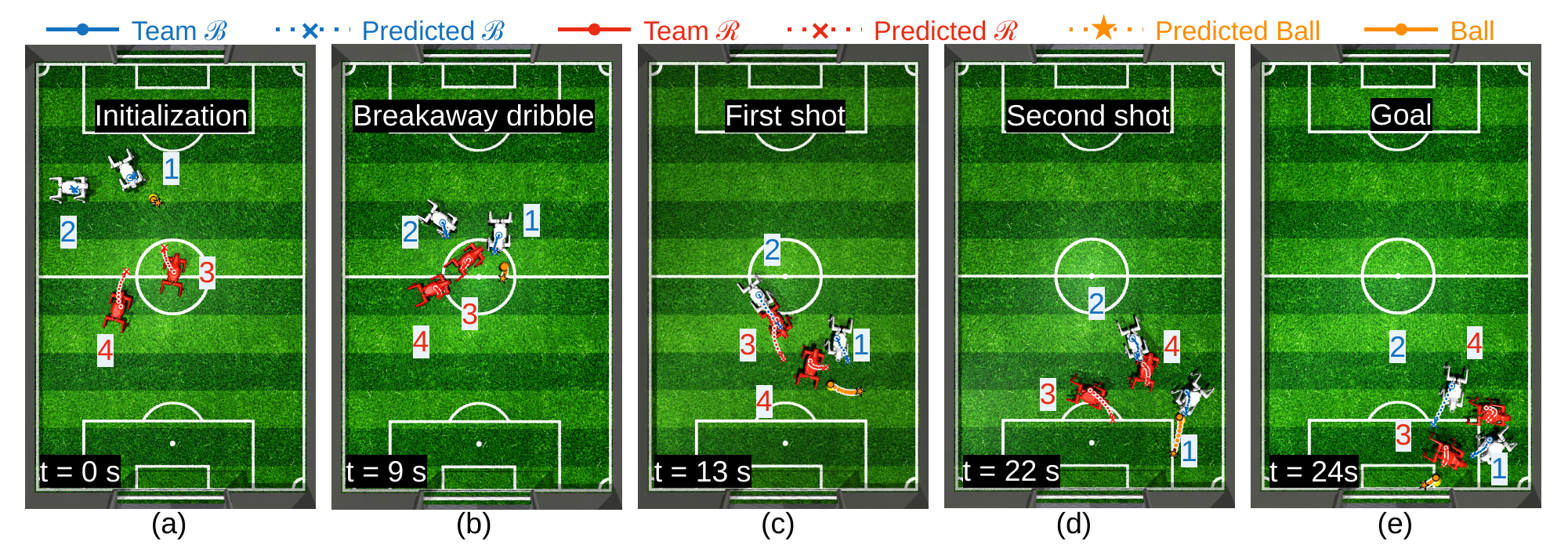}
    \caption{An attacking sequence with simulator snapshots and corresponding MPC-predicted trajectories: (a) initial configuration; (b) robot~1 dribbles past a robot~3; (c) its first shot gets blocked by robot~4, robot~2 repositioning but gets blocked by robot~3; (d) robot~1 attempts a second shot; (e) the ball enters the goal.}
    \label{fig:snapshots}
\end{figure}

In Figure~\ref{fig:snapshots}, we illustrate an attacking sequence involving dribbling, a blocked shot, and a successful second attempt. The MPC trajectories are computed retrospectively from recorded states for visualization and do not control the displayed rollout.

\FloatBarrier
\section{Additional Implementation Details}
\label{apdx:implementation}

\paragraph{State and opponent observations.}
The skill-boundary state contains every robot's base position,
orientation, linear/angular velocity, and joint positions and velocities,
plus ball position/velocity. The residual model predicts all these
continuous quantities at the next high-level boundary; it does not
simulate joint torques or intermediate low-level control steps.
Predicted orientation is projected back to a valid rotation before
constructing local-frame observations. Joint positions are constrained
to simulator joint limits. Agent identity and goal locations are fixed;
the previous skill and command are updated exactly from each branch's
candidate/opponent action, rather than predicted. Actor observations
combine these quantities with relative ball/robot geometry. Thus no
unobserved future simulator state or held-constant joint-state proxy is
used to supply the opponent's proprioceptive inputs. Position-error
tables report robot/ball position components, not a joint-angle error.

\paragraph{Recurrent critic and opponent state.}
Let $h_t^-$ be critic memory immediately before consuming the recorded
state $s_t$. The critic parameters are fixed during collection and its
subsequent teacher queries; the stored memory therefore corresponds to
the same parameter snapshot. Each candidate and ensemble member starts
from an independent copy of $h_t^-$. For every live branch,
\[
\hat h_{t+h}^{+}=\operatorname{GRU}_{\psi}
(\mathcal O_V(\hat s_{t+h}),\hat h_{t+h}^{-}),\qquad
\hat h_{t+h+1}^{-}=\hat h_{t+h}^{+}.
\]
Here $\mathcal O_V$ selects global base/ball poses and velocities.
The recurrence processes $\hat s_t,\ldots,\hat s_{t+H}$ exactly once;
the terminal head reads $\hat h_{t+H}^{+}$. The initial real state is
therefore not consumed twice. Every CEM iteration restarts from the
same copied prefix memory; branches never share updated hidden states.
True termination stops memory propagation and removes the terminal
value. Episode starts alone reset real memory to zero. If a parameter
snapshot is changed before querying, the real episode prefix must be
re-encoded under that snapshot rather than reusing stale memory.
Opponent GRU memory is copied before its observation at $t$ and advanced
with the constructed observation and predicted action at each branch
step, using the frozen rollout opponent. No counterfactual future
observations are taken from the simulator.

\paragraph{Value normalization and reward scale.}
The critic head $v_\psi$ predicts normalized returns. Before GAE or
planning, its output is converted by
\[
V_\psi(\hat s,\hat h^-)=\sigma_R
v_\psi(\hat s,\hat h^-)+\mu_R,\qquad
\sigma_R=\max(\sqrt{\operatorname{Var}(R)},10^{-6}).
\]
The mean/scale are frozen during one collection/query/update cycle;
targets and value clipping (0.2) are expressed in normalized units for
critic optimization. The world-model reward head predicts raw team
rewards, including the same timestep scaling as the simulator. Therefore
both terms in Equation~\ref{eq:mpc_objective} use the same raw discounted
reward scale. The uncertainty coefficient converts dimensionless
state-normalized disagreement to reward units. 
Normalization statistics are updated only between cycles; memory is
recomputed if required by the resulting parameter snapshot.

\paragraph{Scope of the coordination evidence.}
Using the same opponent in collection and planning is a reproducible
design choice, not an independently ablated contribution. Evidence for
coordination transfer is the joint-teacher student versus the independent-
teacher student under identical local observation access, together with
the two event metrics. The no-KL control is a pathway check. Neither
metric-specific ratios nor their bootstrap intervals establish
retention of a general fraction of team coordination ability.

\input{generated/seed_pool_counts.tex}

%% file: generated/learning_controls.tex
\begin{table}[htbp]
\centering\small
\setlength{\tabcolsep}{4pt}
\caption{Fixed-pool all-match win rate (\%) in A40; mean $\pm$ sample SD.}
\label{tab:additional_controls}
\begin{tabular}{lccc}
\toprule
Training variant & 10M & 20M & 40M \\
\midrule
Continuous MAPPO & $40.0\pm12.6$ & $55.0\pm11.5$ & $70.0\pm11.5$ \\
MAPPO + feasibility prior & $44.0\pm12.6$ & $58.0\pm11.5$ & $72.0\pm11.5$ \\
One-step teacher & $50.0\pm12.6$ & $64.0\pm11.5$ & $73.0\pm11.5$ \\
Reward-only teacher & $31.0\pm12.6$ & $44.0\pm11.5$ & $55.0\pm11.5$ \\
Independent MPC teacher & $52.0\pm12.6$ & $65.0\pm11.5$ & $72.0\pm11.5$ \\
Without teacher mask & $57.0\pm12.6$ & $72.0\pm11.5$ & $77.0\pm11.5$ \\
Without distillation & $40.0\pm12.6$ & $55.0\pm11.5$ & $70.0\pm11.5$ \\
Frozen world model & $48.0\pm12.6$ & $60.0\pm11.5$ & $68.0\pm11.5$ \\
Online-only world model & $44.0\pm12.6$ & $57.0\pm11.5$ & $64.0\pm11.5$ \\
Full method & $60.0\pm15.7$ & $76.7\pm13.6$ & $80.0\pm10.5$ \\
\bottomrule
\end{tabular}
\end{table}

%% file: generated/coordination.tex
\begin{table}[htbp]
\centering\small
\setlength{\tabcolsep}{4pt}
\caption{Coordination metrics; mean $\pm$ sample SD across ten paired seeds. Obstruction uses valid-kick events; recovery uses blocked-shot events.}
\label{tab:coordination_controls}
\begin{tabular}{lcc}
\toprule
Controller & Obstruction $\%\downarrow$ & Recovery $\%\uparrow$ \\
\midrule
Frozen-model teacher & $13.0\pm1.3$ & $48.0\pm2.2$ \\
Online-only teacher & $15.0\pm1.5$ & $44.0\pm2.5$ \\
Independent teacher & $12.0\pm1.2$ & $51.0\pm2.0$ \\
Joint teacher & $6.0\pm0.8$ & $64.0\pm1.8$ \\
Joint-teacher student & $7.0\pm0.9$ & $61.0\pm1.9$ \\
Independent-teacher student & $13.0\pm1.3$ & $49.0\pm2.1$ \\
MAPPO student & $18.0\pm1.6$ & $42.0\pm2.4$ \\
\bottomrule
\end{tabular}
\end{table}

%% file: generated/compute.tex
\begin{table}[htbp]
\centering\small
\setlength{\tabcolsep}{4pt}
\caption{Matched interaction and wall-time comparisons. Win rates are all-match percentages; $\pm$ denotes seed SD. Hours are accounting totals on one GPU, not per-seed timing estimates. Unseen W/D/L are seed means.}
\label{tab:compute_controls}
\begin{tabular}{lcccc}
\toprule
Method / budget & Steps (M) & Hours & Fixed-pool W\% & Unseen W/D/L\% \\
\midrule
MAPPO / 40M & 40.0 & 8.0 & $70.0\pm11.5$ & 65.0/8.0/27.0 \\
Full / 40M & 40.0 & 24.2 & $80.0\pm10.5$ & 75.0/8.0/17.0 \\
MAPPO / 24.2 h & 121.0 & 24.2 & $78.0\pm10.5$ & 72.0/8.0/20.0 \\
Full / 8 h & 13.2 & 8.0 & $65.3\pm13.6$ & 60.0/8.0/32.0 \\
\bottomrule
\end{tabular}
\end{table}

%% file: generated/horizon.tex
\begin{table}[htbp]
\centering\small
\setlength{\tabcolsep}{4pt}
\caption{Horizon sweep with fixed candidate count and CEM iterations. Win rates are all-match percentages (mean $\pm$ sample SD); time is the serial-query accounting total.}
\label{tab:horizon_policy}
\begin{tabular}{cccc}
\toprule
Horizon & Win rate at 40M & Hours at 40M & Win rate at 24.2 h \\
\midrule
1 & $72.0\pm11.5$ & 13.7 & $76.0\pm12.6$ \\
2 & $77.0\pm11.5$ & 17.2 & $79.0\pm12.6$ \\
4 & $80.0\pm10.5$ & 24.2 & $80.0\pm10.5$ \\
10 & $74.0\pm11.5$ & 45.8 & $70.0\pm12.6$ \\
\bottomrule
\end{tabular}
\end{table}

%% file: generated/seed_pool_counts.tex
\begin{table}[htbp]
\centering\small
\setlength{\tabcolsep}{4pt}
\caption{40M per-seed W/D/L counts}
\label{tab:pool_seed_counts}
\begin{tabular}{rcccc}
\toprule
Seed & Full / fixed & MAPPO / fixed & Full / unseen & MAPPO / unseen \\
\midrule
0 & 860/76/64 & 799/76/125 & 651/76/273 & 584/76/340 \\
1 & 950/50/0 & 568/77/355 & 585/77/338 & 617/77/306 \\
2 & 890/78/32 & 832/78/90 & 684/78/238 & 815/78/107 \\
3 & 710/79/211 & 667/79/254 & 816/79/105 & 485/79/436 \\
4 & 740/80/180 & 601/80/319 & 717/80/203 & 716/80/204 \\
5 & 830/80/90 & 535/80/385 & 618/80/302 & 551/80/369 \\
6 & 650/81/269 & 865/81/54 & 849/81/70 & 749/81/170 \\
7 & 680/82/238 & 634/82/284 & 915/82/3 & 683/82/235 \\
8 & 920/80/0 & 733/83/184 & 783/83/134 & 782/83/135 \\
9 & 770/84/146 & 766/84/150 & 882/84/34 & 518/84/398 \\
\bottomrule
\end{tabular}
\end{table}

%% file: iclr2027_conference.bib
@inproceedings{10160325,
  author    = {Ji, Yandong and Margolis, Gabriel B. and Agrawal, Pulkit},
  title     = {{DribbleBot}: Dynamic Legged Manipulation in the Wild},
  booktitle = {2023 IEEE International Conference on Robotics and Automation (ICRA)},
  pages     = {5155--5162},
  year      = {2023},
}

@article{chen1998quasi,
  title={A quasi-infinite horizon nonlinear model predictive control scheme with guaranteed stability},
  author={Chen, Hong and Allg{\"o}wer, Frank},
  journal={Automatica},
  volume={34},
  number={10},
  pages={1205--1217},
  year={1998},
  publisher={Elsevier}
}

@inproceedings{pmlrv305su25a,
  author    = {Su, Zhi and Gao, Yuman and Lukas, Emily and Li, Yunfei and Cai, Jiaze and Tulbah, Faris and Gao, Fei and Yu, Chao and Li, Zhongyu and Wu, Yi and Sreenath, Koushil},
  title     = {Toward Real-World Cooperative and Competitive Soccer with Quadrupedal Robot Teams},
  booktitle = {Proceedings of the 9th Conference on Robot Learning},
  series    = {Proceedings of Machine Learning Research},
  volume    = {305},
  pages     = {4723--4742},
  publisher = {PMLR},
  year      = {2025}
}

@inproceedings{venugopal2023mabl,
  author    = {Venugopal, Aravind and Milani, Stephanie and Fang, Fei and Ravindran, Balaraman},
  title     = {{MABL}: Bi-Level Latent-Variable World Model for Sample-Efficient Multi-Agent Reinforcement Learning},
  booktitle = {Proceedings of the 23rd International Conference on Autonomous Agents and Multiagent Systems (AAMAS)},
  pages     = {1865--1873},
  year      = {2024},
}

@inproceedings{yu2022surprising,
  author    = {Yu, Chao and Velu, Akash and Vinitsky, Eugene and Gao, Jiaxuan and Wang, Yu and Bayen, Alexandre and Wu, Yi},
  title     = {The Surprising Effectiveness of {PPO} in Cooperative Multi-Agent Games},
  booktitle = {Advances in Neural Information Processing Systems},
  volume    = {35},
  pages     = {24611--24624},
  year      = {2022},
}

@inproceedings{yang2024sample,
  author    = {Yang, Yaodong and Chen, Guangyong and Hao, Jianye and Heng, Pheng-Ann},
  title     = {Sample-Efficient Multiagent Reinforcement Learning with Reset Replay},
  booktitle = {Proceedings of the 41st International Conference on Machine Learning},
  series    = {Proceedings of Machine Learning Research},
  volume    = {235},
  pages     = {55961--55975},
  publisher = {PMLR},
  year      = {2024}
}

@inproceedings{chua2018deep,
  author    = {Chua, Kurtland and Calandra, Roberto and McAllister, Rowan and Levine, Sergey},
  title     = {Deep Reinforcement Learning in a Handful of Trials Using Probabilistic Dynamics Models},
  booktitle = {Advances in Neural Information Processing Systems},
  volume    = {31},
  pages     = {4759--4770},
  year      = {2018}
}

@inproceedings{janner2019trust,
  author    = {Janner, Michael and Fu, Justin and Zhang, Marvin and Levine, Sergey},
  title     = {When to Trust Your Model: Model-Based Policy Optimization},
  booktitle = {Advances in Neural Information Processing Systems},
  volume    = {32},
  pages     = {12498--12509},
  year      = {2019}
}

@inproceedings{krupnik2020multi,
  author    = {Krupnik, Orr and Mordatch, Igor and Tamar, Aviv},
  title     = {Multi-Agent Reinforcement Learning with Multi-Step Generative Models},
  booktitle = {Proceedings of the Conference on Robot Learning},
  series    = {Proceedings of Machine Learning Research},
  volume    = {100},
  pages     = {776--790},
  publisher = {PMLR},
  year      = {2020}
}

@inproceedings{willemsen2021mambpo,
  author    = {Willemsen, Dani{\"e}l and Coppola, Mario and de Croon, Guido C. H. E.},
  title     = {{MAMBPO}: Sample-Efficient Multi-Robot Reinforcement Learning Using Learned World Models},
  booktitle = {2021 IEEE/RSJ International Conference on Intelligent Robots and Systems (IROS)},
  pages     = {5635--5640},
  publisher = {IEEE},
  year      = {2021},
}

@inproceedings{rusu2015policy,
  author    = {Rusu, Andrei A. and Colmenarejo, Sergio Gomez and Gulcehre, Caglar and Desjardins, Guillaume and Kirkpatrick, James and Pascanu, Razvan and Mnih, Volodymyr and Kavukcuoglu, Koray and Hadsell, Raia},
  title     = {Policy Distillation},
  booktitle = {International Conference on Learning Representations},
  year      = {2016}
}

@inproceedings{ross2011reduction,
  author    = {Ross, St{\'e}phane and Gordon, Geoffrey and Bagnell, Drew},
  title     = {A Reduction of Imitation Learning and Structured Prediction to No-Regret Online Learning},
  booktitle = {Proceedings of the 14th International Conference on Artificial Intelligence and Statistics},
  series    = {Proceedings of Machine Learning Research},
  volume    = {15},
  pages     = {627--635},
  publisher = {PMLR},
  year      = {2011}
}

@inproceedings{tirumala2025robot,
  author    = {Tirumala, Dhruva and Wulfmeier, Markus and Moran, Ben and Huang, Sandy and Humplik, Jan and Lever, Guy and Haarnoja, Tuomas and Hasenclever, Leonard and Byravan, Arunkumar and Batchelor, Nathan and Sreendra, Neil and Patel, Kushal and Gwira, Marlon and Nori, Francesco and Riedmiller, Martin and Heess, Nicolas},
  title     = {Learning Robot Soccer from Egocentric Vision with Deep Reinforcement Learning},
  booktitle = {Proceedings of the 8th Conference on Robot Learning},
  series    = {Proceedings of Machine Learning Research},
  volume    = {270},
  pages     = {165--184},
  publisher = {PMLR},
  year      = {2025}
}

@misc{schuck2026learning,
  author        = {Schuck, Martin and Sorokin, Maks and Manni, Simone and Ta, Duy and Schoellig, Angela P. and Hutter, Marco and Le Cleac'h, Simon and Br{\"u}digam, Jan},
  title         = {Learning Loco-Manipulation From {SMPC} Demonstrations With Sparse Offline-to-Online {RL}},
  year          = {2026},
  eprint        = {2608.12063},
  archivePrefix = {arXiv},
  primaryClass  = {cs.RO}
}

@misc{makoviychuk2021isaac,
  author        = {Makoviychuk, Viktor and Wawrzyniak, Lukasz and Guo, Yunrong and Lu, Michelle and Storey, Kier and Macklin, Miles and Hoeller, David and Rudin, Nikita and Allshire, Arthur and Handa, Ankur and State, Gavriel},
  title         = {{Isaac Gym}: High Performance {GPU}-Based Physics Simulation for Robot Learning},
  year          = {2021},
  eprint        = {2108.10470},
  archivePrefix = {arXiv},
  primaryClass  = {cs.RO}
}

@inproceedings{hansen2022tdmpc,
  author    = {Hansen, Nicklas A. and Su, Hao and Wang, Xiaolong},
  title     = {Temporal Difference Learning for Model Predictive Control},
  booktitle = {Proceedings of the 39th International Conference on Machine Learning},
  series    = {Proceedings of Machine Learning Research},
  volume    = {162},
  pages     = {8387--8406},
  publisher = {PMLR},
  year      = {2022}
}

@inproceedings{zhang2021aorpo,
  author    = {Zhang, Weinan and Wang, Xihuai and Shen, Jian and Zhou, Ming},
  title     = {Model-Based Multi-Agent Policy Optimization with Adaptive Opponent-Wise Rollouts},
  booktitle = {Proceedings of the Thirtieth International Joint Conference on Artificial Intelligence},
  pages     = {3384--3391},
  year      = {2021},
}

@article{zhu2026dynamic,
  author  = {Zhu, Dongjie and Yang, Zhuo and Li, Xuesong and Xu, Wenjun and Liu, Qi and Li, Xiang},
  title   = {Dynamic Legged Ball Manipulation on Rugged Terrains With Hierarchical Reinforcement Learning},
  journal = {IEEE Robotics and Automation Letters},
  volume  = {11},
  number  = {6},
  pages   = {6815--6822},
  year    = {2026},
}

@misc{xiong2024mqe,
      title={MQE: Unleashing the Power of Interaction with Multi-agent Quadruped Environment}, 
      author={Ziyan Xiong and Bo Chen and Shiyu Huang and Wei-Wei Tu and Zhaofeng He and Yang Gao},
      year={2024},
      eprint={2403.16015},
      archivePrefix={arXiv},
      primaryClass={cs.RO}
}

@inproceedings{10802022,
  author    = {Hu, Yutong and Wen, Kehan and Yu, Fisher},
  title     = {{DexDribbler}: Learning Dexterous Soccer Manipulation via Dynamic Supervision},
  booktitle = {2024 IEEE/RSJ International Conference on Intelligent Robots and Systems (IROS)},
  pages     = {12910--12917},
  year      = {2024},
}

@article{wen2026mapmbrl,
  author  = {Wen, Ruoqi and Li, Rongpeng and Xu, Xing and Mashhadi, Mahdi Boloursaz and Xiao, Pei and Zhao, Zhifeng},
  title   = {Multi-Agent Uncertainty-Aware Pessimistic Model-Based Reinforcement Learning for Connected Autonomous Vehicles},
  journal = {IEEE Transactions on Mobile Computing},
  pages   = {1--18},
  year    = {2026},
}

@misc{shi2025gawm,
  author        = {Shi, Zifeng and Liu, Meiqin and Zhang, Senlin and Zheng, Ronghao and Dong, Shanling and Wei, Ping},
  title         = {{GAWM}: Global-Aware World Model for Multi-Agent Reinforcement Learning},
  year          = {2025},
  eprint        = {2501.10116},
  archivePrefix = {arXiv},
  primaryClass  = {cs.MA}
}

@article{wu2023models,
  author  = {Wu, Zifan and Yu, Chao and Chen, Chen and Hao, Jianye and Zhuo, Hankz Hankui},
  title   = {Models as Agents: Optimizing Multi-Step Predictions of Interactive Local Models in Model-Based Multi-Agent Reinforcement Learning},
  journal = {Proceedings of the AAAI Conference on Artificial Intelligence},
  volume  = {37},
  number  = {9},
  pages   = {10435--10443},
  year    = {2023},
}

@article{moerland2023mbrlsurvey,
  author  = {Moerland, Thomas M. and Broekens, Joost and Plaat, Aske and Jonker, Catholijn M.},
  title   = {Model-Based Reinforcement Learning: A Survey},
  journal = {Foundations and Trends in Machine Learning},
  volume  = {16},
  number  = {1},
  pages   = {1--118},
  year    = {2023},
}

@article{hafner2025dreamer,
  author  = {Hafner, Danijar and Pasukonis, Jurgis and Ba, Jimmy and Lillicrap, Timothy},
  title   = {Mastering Diverse Control Tasks Through World Models},
  journal = {Nature},
  volume  = {640},
  number  = {8059},
  pages   = {647--653},
  year    = {2025},
}

@inproceedings{hansen2024tdmpc2,
  author    = {Hansen, Nicklas and Su, Hao and Wang, Xiaolong},
  title     = {{TD-MPC2}: Scalable, Robust World Models for Continuous Control},
  booktitle = {International Conference on Learning Representations},
  year      = {2024}
}

@article{romero2026actorcritic,
  author  = {Romero, {\'A}ngel and Aljalbout, Elie and Song, Yunlong and Scaramuzza, Davide},
  title   = {Actor--Critic Model Predictive Control: Differentiable Optimization Meets Reinforcement Learning for Agile Flight},
  journal = {IEEE Transactions on Robotics},
  volume  = {42},
  pages   = {673--692},
  year    = {2026},
}

@misc{amiri2026pripg,
  author        = {Amiri, Mohsen and Amiri, Mohsen and Beikmohammadi, Ali and Magn{\'u}sson, Sindri and Hosseinzadeh, Mehdi},
  title         = {{PriPG-RL}: Privileged Planner-Guided Reinforcement Learning for Partially Observable Systems with Anytime-Feasible {MPC}},
  year          = {2026},
  eprint        = {2604.08036},
  archivePrefix = {arXiv},
  primaryClass  = {cs.LG}
}

@article{smit2023football,
  author  = {Smit, Andries and Engelbrecht, Herman A. and Brink, Willie and Pretorius, Arnu},
  title   = {Scaling Multi-Agent Reinforcement Learning to Full 11 versus 11 Simulated Robotic Football},
  journal = {Autonomous Agents and Multi-Agent Systems},
  volume  = {37},
  number  = {1},
  pages   = {20},
  year    = {2023},
}

@inproceedings{lanctot2017psro,
  author    = {Lanctot, Marc and Zambaldi, Vin{\'i}cius and Gruslys, Audrunas and Lazaridou, Angeliki and Tuyls, Karl and P{\'e}rolat, Julien and Silver, David and Graepel, Thore},
  title     = {A Unified Game-Theoretic Approach to Multiagent Reinforcement Learning},
  booktitle = {Advances in Neural Information Processing Systems},
  volume    = {30},
  pages     = {4190--4203},
  year      = {2017}
}

@article{vinyals2019alphastar,
  author  = {Vinyals, Oriol and Babuschkin, Igor and Czarnecki, Wojciech M. and Mathieu, Micha{\"e}l and Dudzik, Andrew and Chung, Junyoung and Choi, David H. and Powell, Richard and Ewalds, Timo and Georgiev, Petko and Oh, Junhyuk and Horgan, Dan and Kroiss, Manuel and Danihelka, Ivo and Huang, Aja and Sifre, Laurent and Cai, Trevor and Agapiou, John P. and Jaderberg, Max and Vezhnevets, Alexander S. and Leblond, R{\'e}mi and Pohlen, Tobias and Dalibard, Valentin and Budden, David and Sulsky, Yury and Molloy, James and Paine, Tom L. and Gulcehre, Caglar and Wang, Ziyu and Pfaff, Tobias and Wu, Yuhuai and Ring, Roman and Yogatama, Dani and W{\"u}nsch, Dario and McKinney, Katrina and Smith, Oliver and Schaul, Tom and Lillicrap, Timothy and Kavukcuoglu, Koray and Hassabis, Demis and Apps, Chris and Silver, David},
  title   = {Grandmaster Level in {StarCraft II} Using Multi-Agent Reinforcement Learning},
  journal = {Nature},
  volume  = {575},
  number  = {7782},
  pages   = {350--354},
  year    = {2019},
}
